\documentclass[journal]{IEEEtai}

\usepackage[colorlinks,urlcolor=blue,linkcolor=blue,citecolor=blue]{hyperref}
\usepackage{color,array}
\usepackage{tabularx}
\usepackage{graphicx}
\usepackage{tikz}
\usetikzlibrary{shapes.geometric, arrows}
\usepackage{amsmath}
\usepackage{setspace}
\usepackage{array}
\usepackage{longtable}
\usepackage[edges]{forest}
\usepackage[a4paper, margin=0.5in]{geometry}
\usepackage[ruled,vlined]{algorithm2e}
\usepackage{pgfplots}
\usepackage{pgfplotstable}
\usepackage{booktabs} 
\usepackage{hyperref}
\usepackage{url}
\usepackage{float} 
\usepackage{xcolor}
\usepackage{array}
\usepackage{multirow}
\usepackage{tikz}
\usetikzlibrary{positioning}
\usepackage{xcolor}
\pgfplotsset{compat=1.18}
\usepackage{listings}
\usepackage{adjustbox}
\usetikzlibrary{positioning, shapes, arrows, calc, fit}
\usepackage{caption}

\usetikzlibrary{shapes, arrows.meta, positioning, backgrounds}
\usepackage{titlesec}
\titlespacing*{\section}{0pt}{10pt}{4pt}
\titlespacing{\subsection}{0pt}{10pt}{4pt}
\titlespacing*{\section}{0pt}{10pt}{4pt}
\titlespacing{\subsubsection}{0pt}{8pt}{5pt}

\begin{document}

\title{Mapping the Mind of an Instruction-based Image Editing using SMILE}

\author{
    Zeinab Dehghani$^{1}$, 
    Koorosh Aslansefat$^{1,*}$, 
    Adil Mehmood Khan$^{1}$, 
    Adín Ramírez Rivera$^{2}$, 
    Franky George$^{1}$, 
    and Muhammad Khalid$^{1}$%
    \thanks{$^{1}$Zeinab Dehghani, Koorosh Aslansefat (*Corresponding Author), Adil Khan, Franky George, and Muhammad Khalid are with the Department of Computer Science, University of Hull, HU6 7RX, Hull, UK (e-mail: k.aslansefat@hull.ac.uk).}%
    \thanks{$^{2}$Adín Ramírez Rivera is with the University of Oslo, NO-0316, Oslo, Norway (e-mail: adin.rivera@email.com).}%
}

\maketitle

\begin{abstract}

Despite recent advancements in instruction-based Image Editing models for generating high-quality images, they are known as black boxes and a significant barrier to transparency and user trust. To solve this issue, we introduce SMILE (Statistical Model-agnostic Interpretability with Local Explanations), a novel model-agnostic for localized interpretability that provides a visual heatmap to clarify the textual elements' influence on image-generating models. We applied our method to various Instruction-based Image Editing models like Instruct-pix2pix, Img2Img-Turbo and Diffusers-Inpaint and showed how our model can improve interpretability and reliability. In addition, we use stability, accuracy, fidelity, and consistency metrics to evaluate our method. These findings indicate the exciting potential of model-agnostic interpretability for reliability and trustworthiness in critical applications such as healthcare and autonomous driving, while encouraging additional investigation into the importance of interpretability in enhancing reliable image editing models.
\end{abstract}

\begin{IEEEImpStatement}
The rising complexity and broad utilization of instruction-based Image Editing models require more transparency and user confidence, particularly in sensitive applications where errors or biases can have significant effects. Our proposed framework, SMILE, addresses this requirement by implementing a model-agnostic interpretability method that offers transparent visual observations of the impact of textual instructions on image production. SMILE develops trust and enables stakeholders to improve their understanding and control of the behavior of these systems. This research improves the reliability of AI-driven equipment in fields such as medical imaging, autonomous vehicles, and creative sectors, establishing a basis for more transparent and reliable AI implementations.
\end{IEEEImpStatement}


\begin{figure*}
    \centering
    \includegraphics[width=1\linewidth]{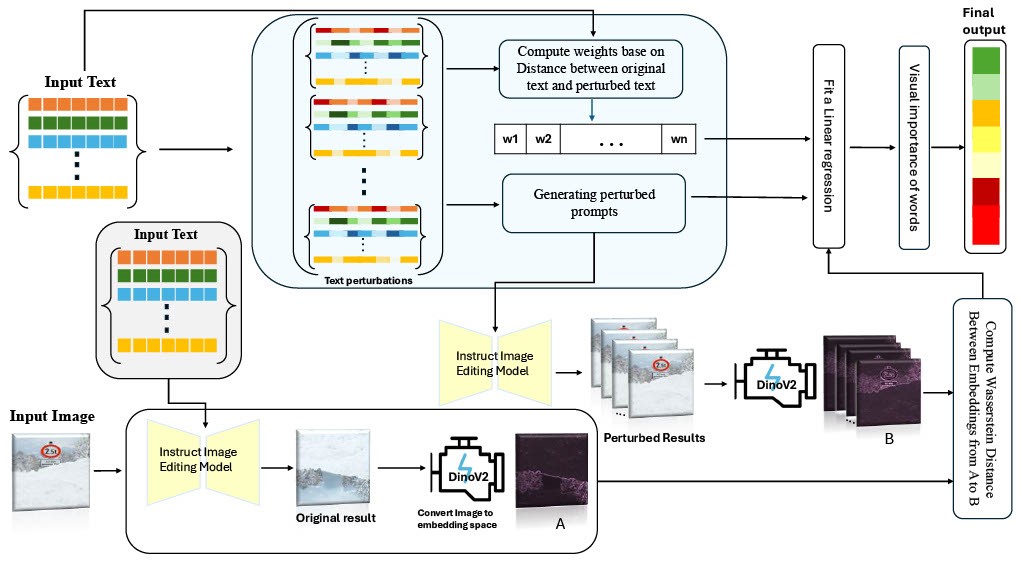}
    \caption{Overview of SMILE for explaining instruct image editing diffusion-based models}
    \label{fig:chart}
\end{figure*}

\section{Introduction}

\IEEEPARstart{T}{he} instruct image editing models have transformed digital content creation by enabling intuitive, text-based modifications to images~\cite{kawar2023imagic}, making complex adjustments accessible across fields such as creative industries~\cite{anantrasirichai2022artificial}, healthcare~\cite{park2020artificial}, marketing~\cite{kopalle2022examining}, education~\cite{osetskyi2020artificial}, and autonomous driving~\cite{huang2020autonomous}.

These tools, built on advanced machine learning models, particularly deep neural networks (DNNs), allow users to make detailed edits using simple natural language commands~\cite{nichol2022glidephotorealisticimagegeneration}. Systems such as Instruct-Pix2Pix~\cite{brooks2023instructpix2pix} use natural language processing (NLP) approaches to match user input with image information. This is achieved using vision-language models, such as CLIP~\cite{radford2021learning}. Lastly, they implement adjustments using generative models such as GANs or Diffusion Models~\cite{goodfellow2020generative, croitoru2023diffusion}.

The problem is that these DNN-based editing systems are still mostly seen as ``black boxes'' and have limited interpretability~\cite{doshi2017towards}. 
This lack of transparency raises significant concerns, particularly in fields where the consequences of model decisions are critical. For example, doctors need to fully understand a model that suggests editing medical images to ensure that critical information is not lost or misrepresented~\cite{holzinger2017we}. Similarly, in autonomous driving, the decisions made by models that process visual input could directly affect safety, which makes it crucial to understand how and why specific edits or interpretations are made~\cite{bojarski2016end}.

The main challenge with these models is their opacity in decision-making. For example, the black-box nature of these models makes it difficult to understand how they interpret and execute commands. For example, when a command like ``remove the pedestrian'' fails, it is unclear whether the model misunderstood the instruction, lacked capability, or encountered another issue~\cite{doshi2017towards}.

This also leads to difficulty debugging, as with insight into the model's internal processes, identifying the source of errors becomes more accessible, making it easier to refine and improve performance effectively~\cite{murdoch2019definitions}. Additionally, there needs to be more efficiency in optimizing commands, as optimizing phrasing becomes a trial-and-error process. Knowing whether to use terms like ``make'' or ``must'' requires extensive testing because it is unclear what the model responds to best~\cite{wen2016network}. Moreover, the model’s hidden internal logic of the model can result in unpredictable performance, causing inconsistency with new or similar commands~\cite{belinkov2019analysis}.

Researchers have developed various methods to address these challenges to improve model explainability. One used approach is LIME (Local Interpretable Model-agnostic Explanations), which approximates the model's behavior locally around a specific prediction using a more straightforward, interpretable model~\cite{ribeiro2016should}. This method helps users determine which input features, like specific parts of an image, had the most significant impact on the model's decision.

BayLIME takes things further by building on LIME and adding Bayesian inference. This means it offers multiple possible explanations while considering uncertainty in the model's predictions~\cite{zhao2021baylime}. It is an upgrade that makes the whole interpretability process even more dependable.

Another popular technique is SHAP (Shapley et al.), which uses a game-theoretic framework to assign each feature a ``Shapley value.''This value represents the feature's contribution to the final prediction~\cite{lundberg2017unified}. SHAP calculates these values by analyzing all possible combinations of characteristics, ensuring consistent and interpretable explanations~\cite{slack2020fooling}. 

In our work, we propose SMILE (Statistical Model-agnostic Interpretability with Local Explanations)~\cite{aslansefat2023explaining}, a novel interpretability method that builds on LIME by integrating Empirical Cumulative Distribution Function (ECDF) statistical distances, providing greater robustness and improved interpretability. 

SMILE generates visual heatmaps that highlight the influence of each word in a text command on the image editing process. When a model is given both an image and a textual instruction, the heatmap will visually display the weight or importance of each word in driving the specific modifications made to the image. This approach gives users a visual map showing how different text parts influence the model’s editing choices. It is a great way to understand better the connection between the text input and the following visual changes.

As shown in Fig.~\ref{fig:chart}, the SMILE framework provides a structured approach to enhancing transparency in instruction-based image editing models by generating perturbed prompts and visual heatmaps. We apply SMILE~\cite{aslansefat2023explaining} across diffusion models like Instruct-Pix2Pix~\cite{an2023fine}, Img2Img-Turbo~\cite{parmar2024one} Diffusers-Inpaint~\cite{rombach2022high}, and DALL-E~\cite{ramesh2022hierarchical}—to assess each model’s interpretability, highlighting word influence on specific edits and demonstrating their potential for user-friendly, explainable image editing.

To ensure a comprehensive assessment of our method, we employ a range of evaluation metrics, including accuracy, fidelity, stability, and consistency~\cite{ahmadi2024explainability}.

The contributions of our work include:
\begin{itemize}
    \item Novel SMILE-Based Interpretability Method: We introduce a new approach that uses SMILE to generate visual heatmaps, highlighting the impact of individual words in textual commands on the image editing process~\cite{aslansefat2023explaining}.
    \item Enhanced Transparency and Trust: By visually mapping the connection between text inputs and visual changes, our approach promotes transparency and builds trust in instruction-based image editing tools, aiding their application across various fields.

    \item Comprehensive System Evaluation: As far as we know,no Token Left Behind—Explainability-Aided Image Classification and Generation using key evaluation metrics—stability, accuracy, fidelity, and consistency. This evaluation demonstrates our method's reliability and effectiveness, setting a new standard for interpretability in this domain~\cite{ahmadi2024explainability}.
\end{itemize}

\section{Literature review}\label{sec2}
This section explores key ideas and recent advancements in instruction-based image editing and explainable AI (XAI), focusing on how they can improve transparency in these models. The discussion is divided into three main areas:

The first area instructs image editing models that allow users to modify images using natural language commands. Here, we provide an overview of these models, including their core technologies, input types and some information.
The second area, Explainable Artificial Intelligence (XAI), addresses the critical need for interpretability in complex AI methods. This section covers various explainability methods, focusing on LIME(Local Interpretable Model-agnostic Explanation) and recent extensions to LIME~\cite{ribeiro2016should}.

The third area, Explainable Instruction Image Editing, focuses on the unique challenges of making instruction-based image editing models interpretable~\cite{lee2023diffusion, evirgen2024text, tang2022daam}.


\subsection{Instruction-Based Image Editing Diffusion models}

Instruction-based image editing modifies an input image according to specific user instructions~\cite{huang2024smartedit}. This task involves providing an instruction that guides the transformation of the original image into a new design that aligns with the given directive~\cite{brooks2023instructpix2pix}. Text-guided image editing enhances the controllability and accessibility of visual manipulation by allowing users to make changes through natural language commands~\cite{fu2023guiding}. Advanced models use large-scale training to perform precise edits efficiently~\cite{kawar2023imagic, huang2024smartedit}.

For example, Instruct-Pix2Pix~\cite{brooks2023instructpix2pix} uses GANs and diffusion models to make detailed edits based on user input, making it a versatile tool for flexible and specific changes. Similarly, Adobe Firefly expands this capability by working with images, videos, and text to create visually enhanced content, often targeting professional use cases. On the other hand, tools like MagicBrush~\cite{zhang2024magicbrush} focus on applying artistic styles and creative effects to images, guided by simple text instructions.

For example, SmartEdit~\cite{huang2024smartedit}utilizes bidirectional interaction to manage complex edits, while MGIE~\cite{shuai2024survey} is specifically designed for large-scale, high-resolution input. Img2Img-Turbo~\cite{parmar2024one} addresses significant challenges traditional diffusion-based models face, such as slow inference times and dependence on paired datasets. It employs a single-step diffusion process and a streamlined generator network, resulting in faster inference, reduced over-fitting, and improved preservation of the structures in the input images. The model excels in unpaired tasks like scene translation (e.g., day-to-night and weather effects) and paired tasks like Sketch2Photo and Edge2Image. Its efficiency and versatility make it a competitive and reliable framework for diverse GAN-based applications. Diffusers-Inpaint~\cite{rombach2022high} specializes in precise inpainting, maintaining seamlessness in modified regions.

Frameworks like MasaCtrl~\cite{cao2023masactrl} and Imagic~\cite{kawar2023imagic} ensure high-quality, realistic results. MasaCtrl excels in non-rigid transformations, while Imagic enables detailed, text-driven edits. MOECONTROLLER~\cite{li2023moecontroller} integrates global and local edits, and Learnable Regions~\cite{lin2024text} uses bounding boxes for localized adjustments without manual masking. DALL-E~\cite{ramesh2022hierarchical} and OmniGen~\cite{xiao2024omnigen} focus on dynamic, layout-preserving transformations and global/local changes, respectively.

Recent advancements improve efficiency and usability. Innovations reduce diffusion models' computational overhead, enhancing practical applications~\cite{ulhaq2022efficient}. FoI (Focus on Your Instruction) isolates relevant regions for precise multi-instruction editing~\cite{guo2024focus}. At the same time, Human-Centered Generative AI (HGAI) aligns generative models with human intent and ethical standards~\cite{chen2023next, chen2023stepshumancenteredgenerativeai}, promoting accessible, creative solutions.

\subsection{Explainable Artificial Intelligence}

Explainable Artificial Intelligence (XAI) is essential as AI models become more complex and more difficult to interpret, especially in critical fields such as healthcare, finance, and autonomous driving, where understanding decisions is crucial~\cite{samek2017explainable, doshi2017towards, rudin2019stop}. Various XAI methods have been developed to address these issues, which fall into two main categories: intrinsic methods and post hoc methods~\cite{guidotti2018survey}. Intrinsic methods involve the use of inherently interpretable models, such as decision trees, designed to be explainable from the outset. In contrast, post hoc methods are applied to complex ``black-box'' models after training, offering interpretability without altering the original model's structure.

Prominent post hoc methods like SHAP~\cite{scott2017unified}, MAPLE~\cite{plumb2018model} and LIME~\cite{ribeiro2016should} , each providing distinct approaches to interpretability. These methods can be further categorized by scope-global methods aim to explain the overall behavior of a model across the entire dataset. In contrast, local methods explain individual predictions by analyzing the model’s behavior around a specific data point~\cite{ribeiro2016should}. 

SHAP is a widely used global method grounded in game theory~\cite{lundberg2017unified}. It quantifies each feature’s impact on model performance and prediction accuracy by considering all possible subsets of features, both with and without the feature. The SHAP value for each feature is calculated as the average marginal contribution across these subsets, allowing users to identify which features significantly influence the model's output~\cite{strumbelj2010efficient}. 

On the other hand, LIME is a local model-agnostic method designed to explain individual predictions~\cite{ribeiro2016should}. Being model-agnostic means that LIME can be applied to any model type without requiring knowledge of its inner workings, making it Multifaceted across different AI applications. LIME achieves interpretability by fitting a simple, interpretable model around the prediction of interest, thereby approximating the complex model’s behavior within a specific region~\cite{ribeiro2016should}. 

MAPLE combines aspects of both global and local explanations, leveraging random forests for feature selection and a local linear model for interpretation~\cite{plumb2018model}. It uses SILO, a technique that weights each sample based on its similarity to other data points, along with DStump, which selects the most critical features~\cite{plumb2018model}. 

Recently, researchers have worked on improving LIME by adapting it for specific types of data. Researchers have introduced adaptations like PointNet LIME for 3D point cloud data~\cite{levi2024fast}, TS-MULE for time series data~\cite{schlegel2021ts}, Graph LIME for graph-structured data~\cite{huang2022graphlime}, Sound LIME for audio data~\cite{mishra2017local}, and B-LIME for ECG signal data~\cite{abdullah2023b}. In addition, several methods have been developed by modifying LIME's approach by using different surrogate models, adjusting distance parameters, altering sampling techniques, and optimizing LIME to improve its interpretability and accuracy.

\begin{figure}[H]
    \centering
    \begin{tikzpicture}[
        node distance=1.5cm, 
        every node/.style={rectangle, draw, rounded corners, font=\small, align=left},
        main/.style={fill=gray!60, minimum width=7.5cm, text width=7.5cm, minimum height=0.75cm, text centered, font=\Large},
        branch/.style={fill=gray!20, minimum width=5cm, text width=5cm, minimum height=0.75cm, text centered, font=\large},
        subbranch/.style={fill=white, minimum width=1.75cm, text width=1.75cm, text centered, font=\small, xshift=0cm},
        description/.style={fill=white, minimum width=6, text width=6cm, font=\small, align=left}
    ]

    \node[main] (root) {Changes in LIME};

    \node[branch, below=0.2cm of root, xshift=-0.75cm] (surrogate) {Use Different Surrogate Model};

    \node[subbranch, below=0.6 of surrogate, xshift=-2cm] (qlime) {Q-LIME};
    \node[description, right=0.2cm of qlime] {Uses a quadratic surrogate model instead of a linear one to better capture non-linear relationships~\cite{bramhall2020qlime}};
    \node[subbranch, below=0.9cm of qlime] (slime) {S-LIME};
    \node[description, right=0.2cm of slime] {Focuses on creating sparse, interpretable explanations by selecting the most relevant features~\cite{zhou2021s}};
    \node[subbranch, below=0.7cm of slime] (bayLime2) {Bay-LIME};
    \node[description, right=0.2cm of bayLime2] {The surrogate model is extended to be a Bayesian linear model~\cite{zhao2021baylime}};
    \node[subbranch, below=0.5cm of bayLime2] (alime) {ALIME};
    \node[description, right=0.2cm of alime] {Employs autoencoders to learn a compressed representation of the input data~\cite{shankaranarayana2019alime}};

    \node[branch, below=0.5cm of alime, xshift=2cm] (distance) {Change Distance Parameter};

    \node[subbranch, below=0.6cm of distance,xshift=-2cm ] (smile) {SMILE};
    \node[description, right=0.2cm of smile] {Uses statistical distance measures (Wasserstein distance instead of Cosine distance)~\cite{aslansefat2023explaining}};

    \node[branch, below=0.5cm of smile, xshift=2cm] (sampling) {Change Sampling Technique};

    \node[subbranch, below=0.6cm of sampling, xshift=-2cm] (slime2) {S-LIME};
    \node[description, right=0.3cm of slime2] {Uses hypothesis testing and the Central Limit Theorem to determine the needed perturbation points~\cite{upadhyay2021extending}};
    \node[subbranch, below=0.8cm of slime2] (anchor) {Anchor};
    \node[description, right=0.3cm of anchor] {Identifies key features through coefficients matching the function’s gradient~\cite{garreau2020explaining}};
    \node[subbranch, below=0.7cm of anchor] (uslime) {US-LIME};
    \node[description, right=0.3cm of uslime] {Chooses data samples close to the decision boundary and near the original data point~\cite{saadatfar2024us}};
    \node[subbranch, below=0.5cm of uslime] (guidedlime) {Guided-LIME};
    \node[description, right=0.3cm of guidedlime] {Uses FCA for structured sampling of instances~\cite{sangroya2020guided}};
    \node[subbranch, below=0.4cm of guidedlime] (dlime) {DLIME};
    \node[description, right=0.3cm of dlime] {Employs AHC and KNN to identify the relevant cluster~\cite{zafar2021deterministic}};
    \node[subbranch, below=0.7cm of dlime] (lslime) {LS-LIME};
    \node[description, right=0.3cm of lslime] {Focuses sampling on relevant parts of the decision boundary rather than the prediction itself~\cite{laugel2018defining}};

    \node[branch, below=0.6cm of lslime, xshift=2cm] (optimize) {Optimize LIME};

    \node[subbranch, below=0.6cm of optimize, xshift=-2cm] (optlime) {OptiLIME};
    \node[description, right=0.3cm of optlime] {Balances explanation stability and model fidelity using mathematical methods~\cite{visani2020optilime}};
    \node[subbranch, below=0.6cm of optlime] (glime) {G-LIME};
    \node[description, right=0.3cm of glime] {Incorporates global context and advanced techniques such as ElasticNet and LARS~\cite{li2023g}};

    \end{tikzpicture}
    \caption{Overview of Variations and Enhancements in the LIME (Local Interpretable Model-Agnostic Explanations) Framework. This figure illustrates key modifications to LIME across different aspects: surrogate models, distance metrics, sampling techniques, optimization methods, and adaptations for specific data types.}
    \label{fig:lime_changes}
\end{figure}
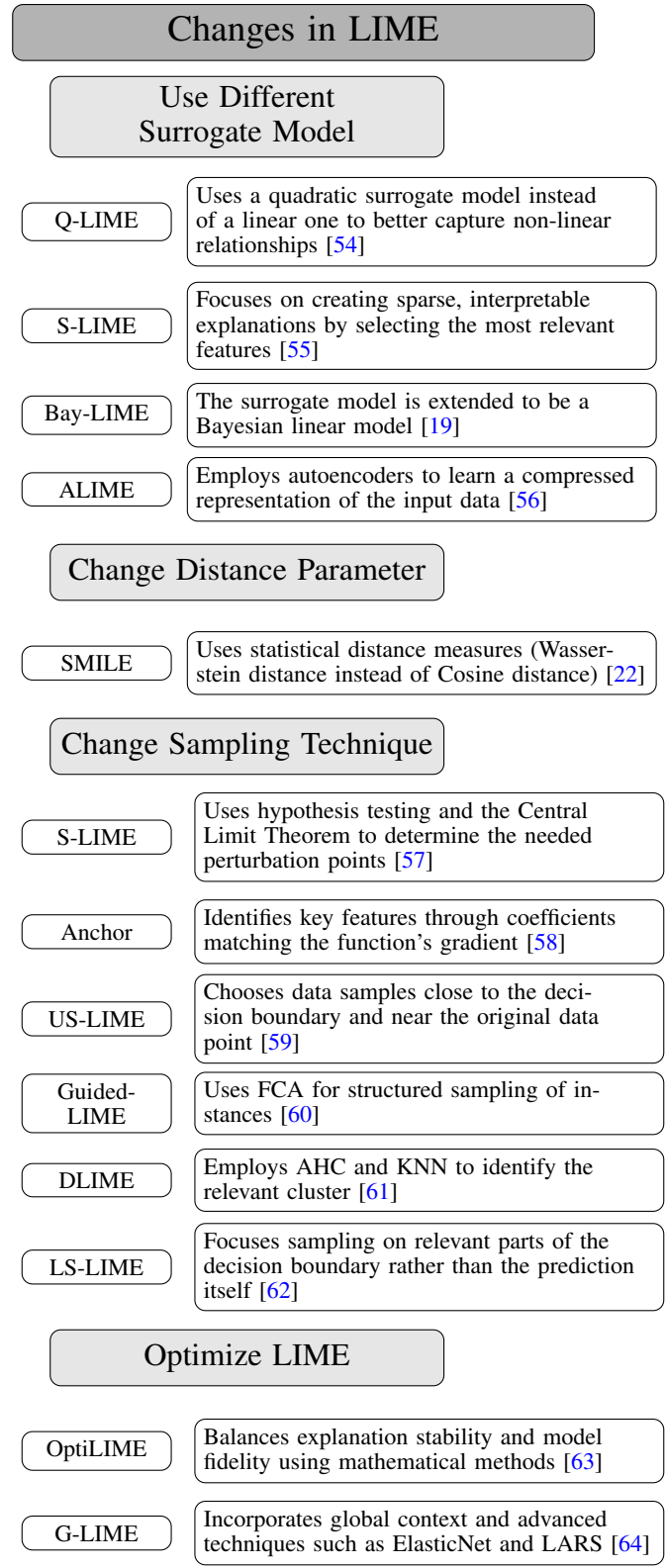

We illustrate some of these methods in Fig.~\ref{fig:lime_changes}.

\subsection{Explainable diffusion based image editing}

Instruction Image Editing models present unique challenges when it comes to explainability. These models convert textual descriptions into images involving complex, high-dimensional transformations. Consequently, the internal workings of these models could be more inherently interpretable. For instance, identifying which parts of the input text influence specific regions of the generated image remains challenging~\cite{lee2023diffusion, evirgen2024text, tang2022daam, dang2024explainable}.

Previous studies employed local XAI methods like heatmaps to visually map text features to corresponding image regions. DAAM~\cite{tang2022daam} introduces an innovative enhancement for capturing self-attention within images, and the text-image cross-attention initially addressed. By guiding self-attention, DAAM-I2I~\cite{chowdhury2024daam} significantly improves heatmap quality, leading to more precise localization in segmentation and object detection tasks.

Recent research on robustness, fairness, security, privacy, factuality, and explainability addresses ethical, harmful outputs and social concerns~\cite{zhang2024trustworthy, hao2024harm}. 

Evirgen et al.~\cite{evirgen2024text} tackled user interaction challenges with text-to-image models by emphasizing using example-based explanations and curating datasets to better tailor explanations for beginners. Similarly, PromptCharm~\cite{wang2024promptcharm} introduced model explanations by visualizing attention values, allowing users to refine their prompts and produce higher-quality outputs interactively.

Patcher~\cite{chang2024repairing} introduced a method for mitigating catastrophic neglect in text-to-image models through attention-guided feature enhancement. This approach improves the alignment between prompts and generated content, directly addressing semantic inconsistencies. In addition to text-to-image generation, diffusion-based models have been applied to tasks such as open-vocabulary segmentation~\cite{karazija2023diffusion, karazija2025diffusion} and person detection dataset generation~\cite{rodriguez2024exploring}.

OVDiff~\cite{karazija2023diffusion} leverages pre-trained diffusion models for zero-shot segmentation without additional training. FOSSIL~\cite{barsellotti2024fossil} integrates text-conditioned diffusion models and self-supervised features for unsupervised segmentation, capturing semantic variability while improving explainability.

In video generation, Vico~\cite{yang2024compositional, yang2024compositionalvideogenerationflow} analyzes the token influence and balances latent updates, enhancing compositional accuracy in video generation. Similarly, MultiLate~\cite{vetagiri2024multilate} employed attention attribution maps to enhance multimodal hate speech detection, showcasing explainability's impact in improving classification performance.

Methods for generating controllable RGBA illustrations~\cite{quattrini2024alfie} and fine-grained visuo-spatial representations for embodied AI tasks~\cite{gupta2024pre} showcase diffusion models' potential in creative workflows. Finally, Tankelevitch et al.~\cite{tankelevitch2024metacognitive} proposed integrating explainability to reduce cognitive load in generative workflows.

These contributions highlight the importance of explainability in improving the usability, trustworthiness, and transparency of diffusion-based generative AI systems across diverse applications.

In this work, we introduce SMILE (Statistical Model-agnostic Interpretability with Local Explanations) to create heatmaps that show how specific text elements influence image edits in instruction-based editing~\cite{aslansefat2023explaining}. SMILE, which is model-agnostic, is chosen for its compatibility with various models~\cite{ribeiro2016should}, drawing from methods similar to LIME but with improved robustness~\cite{gilpin2019explainingexplanationsoverviewinterpretability}. Unlike LIME, which can be vulnerable to adversarial attacks that distort interpretability, SMILE is more resistant to manipulation~\cite{aslansefat2023explaining}. These heatmaps make interactions with editing models more understandable and reliable, helping users trust the edits by offering insights that are less prone to distortion.

\section{Problem Definition}
In complex deep learning models, extensive language and text-based image editing models, interpretability poses a significant challenge. Due to the Enormous number of parameters and complex processing mechanisms, these models essentially act as black boxes, making it problematic for users to understand why and how the model responds to certain inputs or shows specific behaviors~\cite{doshi2017towards, rudin2019stop}. The need for interpretability poses a severe challenge to these models' safe deployment and control. It makes it harder to manage them effectively and increases the risk of unexpected or undesirable behavior.

For instance, to improve the interpretability of large language models, the Anthropic team applied Sparse Autoencoders to extract semantic and abstract features from these models. This technique enabled them to identify critical concepts and internal behaviors, thus providing a more transparent view of how these models operate~\cite{templeton2024scaling}.

Inspired by this approach and recognizing the importance of interpretability in text-based image editing models, we present an innovative, model-agnostic approach for evaluating and enhancing the interpretability of these models. We leverage SMILE (Statistical et al. with Local Explanations) as a foundational tool to generate localized explanations and visual heatmaps. These heatmaps  illustrate the influence of specific text elements on the image generation process, helping users better understand the relationship between textual inputs and visual outputs~\cite{aslansefat2023explaining}. This model-agnostic method is crucial for making instruction image editing models more transparent and trustworthy.

\begin{figure}[H]
    \centering
    \includegraphics[width=1\linewidth]{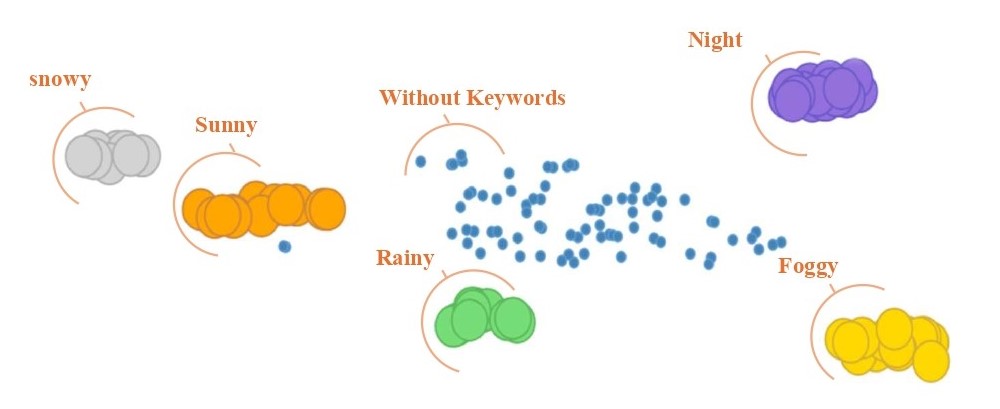}
    \caption{2D t-SNE visualization of image embeddings, showing distinct clusters for different prompt keywords, with clear separability between groups representing specific text features and those with perturbed, non-specific prompts.}
    \label{fig:tsne_5}
\end{figure}

In addition, to further examine the model’s interpretability, We incorporated a t-SNE scatter plot, as you can see in Fig.~\ref{fig:tsne_5} to visualize the embeddings of images modified with various descriptive sentences, each embedding color-coded based on specific keywords. In this plot, images generated with sentences containing the keyword ``sunny'' are represented as orange points, ``rainy'' as green points, ``foggy`` as yellow points, ``snowy'' as gray points, and ``night'' as purple points. Additional points generated from sentences with perturbed texts lacking targeted keywords are represented as dark blue points.

This visualization shows that embeddings of each keyword  create discrete clusters, clearly separated from perturbation points without specific keywords. This separation suggests that a surrogate model, like weighted linear regression, could effectively capture the model's behavior. Moreover, the regression coefficients provide a simple, interpretable measure of each keyword's importance, showing how they influence the model's output. This method enhances interpretability by offering a clear, quantitative view of how specific textual components impact the generated images.

 \section{Proposed Method} \label{sec3}
 
The proposed method improves the interpretability of instruction image editing models by focusing on understanding how specific text inputs influence the generated images~\cite{brooks2023instructpix2pix}. Analyzing the relationship between textual prompts and visual output makes the behavior of the model more transparent and predictable~\cite{fu2023guiding}.

\begin{figure}[H]
    \centering
    \includegraphics[width=1.2\linewidth]{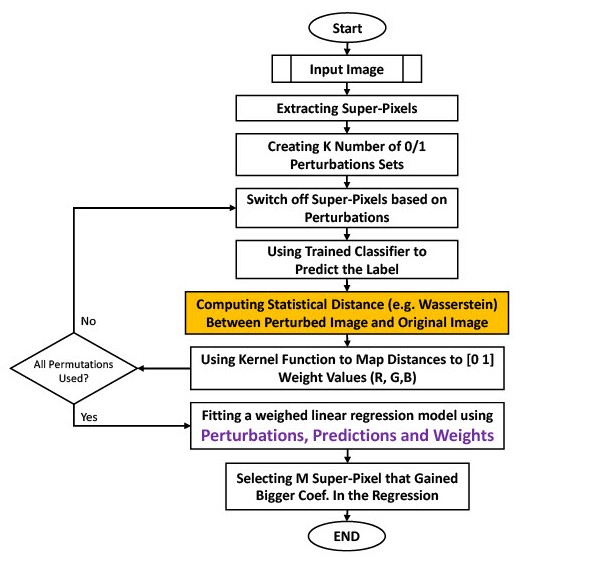}
    \caption{SMILE flowchart for explaining image classification \cite{aslansefat2023explaining}}
    \label{fig:smile}
\end{figure}

In the context of image classification, as you can see in Fig.~\ref{fig:smile}, SMILE is a tool that explains why a model makes confident predictions by isolating specific parts of an image that influence the decision. SMILE starts by segmenting the image into ``super-pixels,'' cohesive clusters of pixels representing distinct image parts. It then creates multiple altered versions of the image by randomly keeping or removing different superpixels as perturbations, generating images highlighting different regions of the original.

For each of these perturbed images, SMILE records the model's predictions. It calculates a similarity score with the original image using a statistical measure based on the Empirical Cumulative Distribution Function (ECDF), such as the Wasserstein distance. These similarity scores are then mapped to weights using a kernel function, which transforms the distances to fall within a specified range, allowing for a balanced weighting scheme.

Finally, SMILE uses the perturbations, predictions, and computed weights to train a weighted linear regression model. This model identifies the superpixels that most strongly influence the model's final decision, pinpointing specific shapes, colors, or textures as key factors in the prediction. The resulting explanation clearly shows which parts of the image were vital in the model's prediction. Inspired by SMILE's approach, we adapted this methodology to interpret how text prompts influence image generation in our work.

\begin{figure*}
    \centering
    \includegraphics[width=1\linewidth]{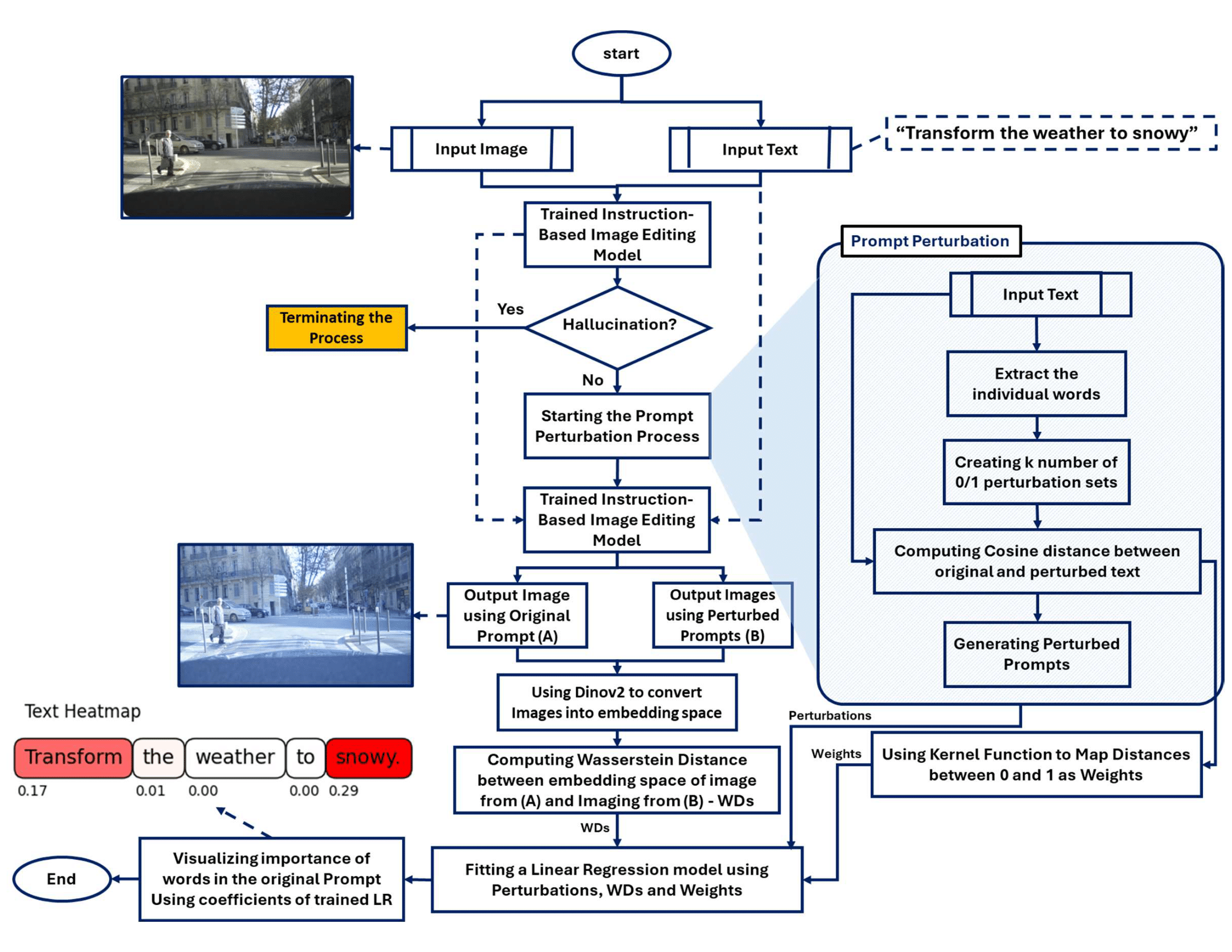}
    \caption{SMILE flowchart for explaining instruct based image editing models}
    \label{fig:flow}
\end{figure*}

Inspired by SMILE's capabilities, we propose a similar method for interpreting text-prompt-based image generation models as demonstrated in Fig.~\ref{fig:flow}. Instead of perturbing the generated image, as done in SMILE, we modify the instruction by including or excluding certain words. We generate a corresponding image for each altered text prompt and compute the Wasserstein distance between these images and the one produced by the original text prompt. This distance is a similarity measure, allowing us to identify which words in the text prompt have the most significant impact on the generated image.

These similarity scores are then used as the outcome variable in a weighted linear regression model, with weights based on the text distances and perturbations. This regression model helps determine how each word affects the finished image. The coefficients derived from this model quantify the influence of each word, which we use to create a visual heatmap highlighting the most influential words in the text prompt. This heatmap provides an intuitive visual representation of which prompt elements contribute most to the generated output.

By employing this enhanced three-step process that builds on SMILE's methodology, we aim to help users understand how particular elements of text prompts impact image generation~\cite{huang2024smartedit}. This technique improves user control and predictability in instruction-driven image editing models by offering insightful information about which elements of textual input—such as stylistic instructions or descriptive terms—impact the final visual outputs most. 

The first step employs an image editing model based on input text. The image editing model receives the original image input and an corresponding text; only the input text changes each time the model renders the image. The image editing model is expected to generate images similar to the original image that vary only based on critical components of the input texts. 

Different text permutations are utilized from the individual words of the input prompt to create various text perturbation blocks after breaking down the prompt into individual words~\cite{liu2023toward}. The instruction-based image editing model generates images based on the original input and the perturbation text. 

While the Instruction-Based Image Editing model faces challenges in counting objects and spatial reasoning, these issues can be mitigated by carefully selecting perturbation texts and restricting the input domain for perturbations~\cite{qiu2023benchmarking}. Instruction-based image editing helps the proposed method create images that only conflict with the original text regarding keywords related to perturbation texts.

Second, To assess similarities between original and modified images, we use the DINOv2 model to generate high-quality, semantic embeddings~\cite{oquab2023dinov2}. Direct image comparisons often capture only surface features like color or pixel structure, overlooking more profound semantic shifts. Embedding images in DINOv2’s space enables us to capture visual and conceptual content, allowing for more meaningful comparisons~\cite{caron2021emerging}.

This project aims to develop a model-agnostic, explainable framework for interpreting image edits independent of specific models, such as Instruct-Pix2Pix~\cite{an2023fine} or DALL-E~\cite{ramesh2022hierarchical}. DINOv2 is particularly suitable for this purpose due to its self-supervised learning approach, which enables it to generate high-quality, general-purpose image embeddings without relying on labeled data or model-specific outputs. Its adaptability to diverse image types makes it ideal for model-agnostic explainability~\cite{oquab2023dinov2, caron2021emerging}.

We use the Wasserstein Distance to quantify these differences to compare the distributions in both image and text embeddings~\cite{rubner2000earth, arjovsky2017wasserstein}. We use the Wasserstein distance because it provides a more complete measure of the differences between image embeddings compared to other ECDF-based distances like Kolmogorov-Smirnov or Cramér–von Mises. While those methods focus on specific aspects, Wasserstein distance captures both broad and subtle changes in the data~\cite{rubner2000earth}. 

Wasserstein Distance is also effective for text embeddings by capturing distributional geometry, enabling us to assess how text perturbations conceptually align with the original prompt~\cite{kusner2015word}.

Finally, by applying linear regression, the space of the input prompts is linked to the image-generating space of the Instruction-Based Image Editing model. As a result, the Instruction-Based Image Editing can be interpreted by examining the regression coefficients~\cite{murdoch2019interpretable}.

\subsection{Image Generating and perturbed prompts}

To begin, we generate variations of the original text prompt, known as ``perturbed prompts'' to examine how these subtle changes affect the output of an image-editing model~\cite{fu2023guiding}. The original text is first broken down into individual words. Multiple text versions are then created by selectively including or excluding words. Each subset is paired with an input image and provided to the Instruct Image Editing model, which generates a unique image for each prompt. In other words, the input image is repeatedly edited with different text prompts~\cite{huang2024smartedit}.In this process, the image edited with the original sentence is considered the reference, or baseline image, against which all other edited images will be compared.

For the original text prompt \( p_{\text{org}} \), the Instruct Image Editing function \( \phi_{\text{edit}} \) generates an output image \( \alpha_{\text{org}}' \) based on the input image  \( \alpha \), as shown in Eq.~\ref{eq:original_image_generation}:

\begin{equation}
\label{eq:original_image_generation}
\alpha_{\text{org}}' = \phi_{\text{edit}}(p_{\text{org}}, \alpha_{\text{input}})
\end{equation}

Similarly, for each perturbation text \( p_{\text{pert}}^{i} \) within the set \( S_{\text{pert}} \), the network produces a corresponding edited image \( \alpha_{\text{pert}}'^{i} \) according to Eq.~\ref{eq:perturbed_image_generation}:

\begin{equation}
\label{eq:perturbed_image_generation}
\alpha_{\text{pert}}'^{i} = \phi_{\text{edit}}(p_{\text{pert}}^{i}, \alpha_{\text{input}}), \quad \forall i \in S_{\text{pert}}.
\end{equation}

Here, \( p_{\text{org}} \) represents the original prompt, \( p_{\text{pert}}^{i} \) describes a perturbation generated from the original text, and \( \alpha_{\text{input}} \) is the input image used by the Instruct Image Editing model function, \( \phi_{\text{edit}} \), to generate images \( \alpha' \) that reveal the influence of the prompt variations. The set of all perturbation texts is denoted by \( S_{\text{pert}} \)~\cite{ribeiro2016should, brooks2023instructpix2pix}.

\subsection{Creating the interpretable space}

In order to map the output image space of the Instruct Image Editing model to a one-dimensional space, the Wasserstein distance between each image generated from the perturbed texts and the original text(baseline image) is computed as a measure of similarity~\cite{arjovsky2017wasserstein}. In all cases, the input image for editing remains consistent. A self-supervised feature extraction model, DINOv2, is employed to eliminate redundant information from the image, allowing focus on the primary components of the image~\cite{oquab2023dinov2, peyre2019computational}. This reduces noise in the distance estimation, enabling the accurately calculated distance to serve as the output for the linear regression model.

The distance between the images generated by the perturbation texts and the original text is given by Eq.~\ref{eq:eq3}:
\begin{equation}
\label{eq:eq3}
\begin{aligned}
W_i(\alpha_{\text{org}}', \alpha_{\text{prom}}'^{i}) &= \left( \frac{1}{n} \sum_{j=1}^{n} \left\| \delta_{\text{Dinov2}}(\alpha_{\text{org}}'^{j}) \right. \right. \\
& \left. \left. \quad - \, \delta_{\text{Dinov2}}(\alpha_{\text{prom}}'^{i,j}) \right\|^p \right)^{\frac{1}{p}}, \\
& \forall i \in S_{\text{prom}}
\end{aligned}
\end{equation}

In the above relation, the \( \delta_{\text{Dinov2}} \) function represents the feature extraction model applied to the images before calculating the distance~\cite{oquab2023dinov2}. In Eq.~\ref{eq:eq3}, \( n \) represents the number of pixels, \( p \) denotes the norm order, and \( W(\alpha_{\text{org}}, \alpha_{\text{pert}}'^{i}) \) demonstrates the Wasserstein distance between the two images produced by the original and perturbed prompts~\cite{gulrajani2017improved}.

Using p-values is crucial for validating the significance of the observed Wasserstein distances (\texttt{WD}) between images generated under different prompt perturbations. By calculating a p-value, we test whether the observed \texttt{WD} could have occurred by chance, determining if the perturbation-induced changes are meaningful~\cite{wasserman2013all}. To compute this p-value, we use the Bootstrap algorithm~\cite{tibshirani1993introduction, gilleland2020bootstrap}, which estimates the probability of observing a \texttt{WD} as large as the observed one. This involves generating distributions of \texttt{WD}s through repeated sampling and comparing each sampled \texttt{WD} with the observed value. If a small proportion of sampled \texttt{WD}s exceed the observed \texttt{WD}, the resulting p-value confirms that the differences are significant, supporting the validity of the perturbation effects.

The \textbf{Wasserstein-\( p \) metric} allows us to adjust \texttt{WD}’s sensitivity to different image variations by fine-tuning the norm order \( p \). Smaller  \( p \) values emphasize local differences, while larger  \( p \) values capture broader distributional shifts~\cite{peyre2019computational}. By testing various  \( p \)-values, we identify the norm that best captures meaningful variations caused by prompt changes, enhancing the reliability and interpretability of the results~\cite{peyre2019computational}.

When analyzing Wasserstein distances, measures with high p-values—indicating non-significant differences—are excluded from the procedure~\cite{gulrajani2017improved}. The Bootstrap algorithm, described in Algorithm~\ref{alg:bootstrap}, calculates these p-values for all relevant ECDF-based distance measures, including \texttt{WD}. The algorithm computes both the \texttt{WD} and its associated p-value.

For a univariate example, let \texttt{X} and \texttt{Y} be the inputs. The Bootstrap algorithm performs \texttt{1e5} iterations, where \texttt{XY} is the concatenated set of \texttt{X} and \texttt{Y}. In each iteration, two random samples are drawn from \texttt{XY}, and their Wasserstein distance (\texttt{boostWD}) is computed. If \texttt{boostWD} exceeds the observed \texttt{WD}, a counter (\texttt{bigger}) is incremented. After completing all iterations, the p-value (\texttt{pVal}) is calculated as the proportion of iterations where \texttt{boostWD} is greater than the observed \texttt{WD}, as shown in Algorithm~\ref{alg:bootstrap}.

\begin{algorithm}
    \caption{Bootstrap Algorithm for WD P-Value Calculation}
    \label{alg:bootstrap} 
    \SetAlgoLined
    \KwResult{p-value and WD}
    MaxItr = 1e5\;
    WD = Wasserstein\_Dist(X, Y)\;
    XY = Concatenate(X, Y)\;
    LX = len(X)\;
    LY = len(Y)\;
    n = LX + LY\;
    bigger = 0\;
    
    \For{ii in 1 to MaxItr}{
        e = random.sample(range(n), LX)\;
        f = random.sample(range(n), LY)\;
        boostWD = Wasserstein\_Dist(XY[e], XY[f])\;
        \If{boostWD $>$ WD}{
            bigger = 1 + bigger\;
       } 
   } 
    pVal = bigger / MaxItr\;
    \Return pVal, WD\;
\end{algorithm}

When generating perturbation texts derived from the original text, the keywords of the original text play a crucial role in maintaining certain levels of similarity and fidelity between the perturbation texts and the source content. This can be important when the interpretable model is trained on perturbation texts as input. Thus, utilizing a similarity criterion, the degree of similarity of each perturbation text with the original text is measured and employed as a sample weight for each text in the interpretable model. Eq.~\ref{eq:eq4} calculates the similarity of perturbation texts with the original text, and the Gaussian kernel is applied to be in the range of zero to one~\cite{kusner2015word}.

\begin{equation}
\begin{aligned}
\text{WMD}(p_{\text{org}}, p_{\text{prom}}^{i}) &= \min_{T \geq 0} \sum_{k=1}^{n} \sum_{l=1}^{m} T_{kl} \, d(w_k, w_l) \\
\text{subject to} \quad & \sum_{l=1}^{m} T_{kl} = p_k, \quad \forall k \in S_{\text{org}} \\
& \sum_{k=1}^{n} T_{kl} = q_l, \quad \forall l \in S_{\text{prom}} \\
& T_{kl} \geq 0
\end{aligned}
\end{equation}

\begin{equation}
\label{eq:eq4}
\begin{aligned}
\pi_i(p_{\text{org}}, p_{\text{prom}}^{i}) &= \exp \left( 
    - \left( 
        \frac{C_i(p_{\text{org}}, p_{\text{prom}}^{i})}{\sigma^2} 
    \right)^2 
\right), \\
& \quad \forall i \in S_{\text{prom}}
\end{aligned}
\end{equation}

The above relationship, \( WMD(p_{\text{org}}, p_{\text{prom}}^{i}) \) indicates the similarity of the perturbation text with the original text, and \( \pi_i(p_{\text{org}}, p_{\text{prom}}^{i}) \) represents the weight of each disturbed text for the regression process, which is obtained by applying the Gaussian kernel~\cite{hastie2009elements}.

\subsection{Developing the Interpretable Surrogate Model}

We are innovative in mapping the output space of the Instruct Image Editing model to a one-dimensional distance scale and accompanying this with corresponding perturbation texts for the calculated distances~\cite{habib2024exploring}. This method enables us to train an interpretable model on the mapped output space of the Instruct Image Editing model, effectively demonstrating how input words influence image generation.

In our linear regression model, the vectors of the perturbation texts are critical, as they are treated as independent variables. Following the extraction of image features, the distance between images of these perturbation texts and the corresponding original text image is regarded as the response variable. Additionally~\cite{radford2021learning}, the degree of similarity between each perturbation text and the original text is assumed to represent the weight of the samples~\cite{molnar2020interpretable}. Eq.~\ref{eq:eq5} outlines the square loss objective function for interpretable weighted linear regression.

\begin{equation}
\label{eq:eq5}
\begin{aligned}
\text{Loss}(\phi_{\text{edit}}, \delta_{\text{Dinov2}}, p_{\text{org}}) =  \quad  \quad \quad \quad \quad \quad \\
\frac{1}{n} \sum_{i=1}^{n} \pi_i(p_{\text{org}}, p_{\text{prom}}^{i}) 
\times \left( W_i(\alpha_{\text{org}}', \alpha_{\text{prom}}'^{i}) - \hat{f}_t(p_{\text{prom}}^{i}) \right)^2
\end{aligned}
\end{equation}


After feature extraction, the \( \hat{f}_i(p_{\text{prom}}^{i,01}) \) function is a linear regression model trained over the one-hot vector space of perturbation texts \( (p_{\text{prom}}^{i,01}) \) and image distances~\cite{molnar2020interpretable}.

\subsection{Evaluation Metrics}

In our investigation, we adopt a suite of evaluation metrics inspired by foundational work provided by Google~\cite{sanchez2020evaluating}, emphasizing the multifaceted nature of assessing explainable models. This work highlights the significance of metrics such as accuracy, stability, fidelity and consistency as essential tools for a rigorous evaluation of model behavior, particularly when comparing explainable models to traditional black-box models. The adoption of these metrics provides a structured methodology to dissect and understand model reliability in a more holistic manner~\cite{sanchez2020evaluating}.

\subsubsection{Accuracy}

Accuracy measures how well the model's output aligns with the expected results (ground truth). Specifically, we compare the model's attribution scores (attention to text elements) against ground truth labels that identify the most relevant text elements for specific image regions~\cite{fong2017interpretable}.

To quantify this, we use the \textbf{Area Under the Curve (AUC)} of the Receiver Operating Characteristic (ROC)~\cite{hanley1982meaning}. AUC reflects the model's ability to rank relevant elements (from the ground truth) higher than irrelevant ones:
\begin{itemize}
    \item \textbf{AUC $\sim$1:} The model effectively distinguishes relevant elements, closely matching human-identified ground truth.
    \item \textbf{AUC $\sim$0.5:} The model's ranking is random, showing poor alignment with the ground truth.
\end{itemize}

\begin{figure}[H]
    \centering
    \begin{tikzpicture}
    \definecolor{whitebox}{rgb}{1, 1, 1}
    \definecolor{lightpink}{rgb}{1, 0.8, 0.8}
    \definecolor{mediumred}{rgb}{1, 0.6, 0.6}
    \definecolor{red}{rgb}{1, 0.4, 0.4}
    \definecolor{darkred}{rgb}{1, 0.2, 0.2}

    \node[align=center, font=\bfseries] at (-2, 5) {Attribution Accuracy Test};
    \node[align=center, font=\bfseries] at (-5, 4.5) {Ground Truth};

    \node[draw, rounded corners, fill=whitebox, minimum width=1.3cm] (t1) at (-5.5, 4) {could};
    \node[draw, rounded corners, fill=whitebox, minimum width=1.3cm] (t2) at (-4.2, 4) {you};
    \node[draw, rounded corners, fill=whitebox, minimum width=1.3cm] (t3) at (-2.9, 4) {please};
    \node[draw, rounded corners, fill=darkred, minimum width=1.3cm] (t4) at (-1.6, 4) {make};
    \node[draw, rounded corners, fill=whitebox, minimum width=1.3cm] (t5) at (-0.3, 4) {this};
    \node[draw, rounded corners, fill=darkred, minimum width=1.3cm] (t6) at (1, 4) {rainy};

    \node[draw, rounded corners, fill=blue!20, minimum width=2.5cm, minimum height=0.5cm] (pix2pix1) at (-2.3, 2.5) {Generative Model};

    \node[align=center, font=\bfseries] at (-5, 2) {Model 1: AttAUC = 0.8};

    \node[draw, rounded corners, fill=lightpink, minimum width=1.3cm] (w1) at (-5.5, 1.5) {could};
    \node[draw, rounded corners, fill=darkred, minimum width=1.3cm] (w2) at (-4.2, 1.5) {you};
    \node[draw, rounded corners, fill=whitebox, minimum width=1.3cm] (w3) at (-2.9, 1.5) {please};
    \node[draw, rounded corners, fill=darkred, minimum width=1.3cm] (w4) at (-1.6, 1.5) {make};
    \node[draw, rounded corners, fill=red, minimum width=1.3cm] (w5) at (-0.3, 1.5) {this};
    \node[draw, rounded corners, fill=darkred, minimum width=1.3cm] (w6) at (1, 1.5) {rainy};

    \node[align=center, font=\bfseries] at (-5, 0.3) {Model 2: AttAUC = 1};

    \node[draw, rounded corners, fill=lightpink, minimum width=1.3cm] (j1) at (-5.5, -0.2) {could};
    \node[draw, rounded corners, fill=lightpink, minimum width=1.3cm] (j2) at (-4.2, -0.2) {you};
    \node[draw, rounded corners, fill=whitebox, minimum width=1.3cm] (j3) at (-2.9, -0.2) {please};
    \node[draw, rounded corners, fill=darkred, minimum width=1.3cm] (j4) at (-1.6, -0.2) {make};
    \node[draw, rounded corners, fill=mediumred, minimum width=1.3cm] (j5) at (-0.3, -0.2) {this};
    \node[draw, rounded corners, fill=darkred, minimum width=1.3cm] (j6) at (1, -0.2) {rainy};

    \node at (-5.5, 3.5) {0};
    \node at (-4.2, 3.5) {0};
    \node at (-2.9, 3.5) {0};
    \node at (-1.6, 3.5) {1};
    \node at (-0.3, 3.5) {0};
    \node at (1, 3.5) {0};

    \node at (-5.5, 1) {0.10};
    \node at (-4.2, 1) {0.80};
    \node at (-2.9, 1) {0.01};
    \node at (-1.6, 1) {0.70};
    \node at (-0.3, 1) {0.50};
    \node at (1, 1) {0.90};

    \node at (-5.5, -0.7) {0.10};
    \node at (-4.2, -0.7) {0.10};
    \node at (-2.9, -0.7) {0.01};
    \node at (-1.6, -0.7) {0.70};
    \node at (-0.3, -0.7) {0.20};
    \node at (1, -0.7) {0.90};

    \end{tikzpicture}
    \caption{Accuracy evaluation using AttAUC (Attention Area Under Curve) metric. The ground truth highlights `make'' and rainy'' as the most influential words for generating the desired image. The model’s predicted attention scores are compared to the ground truth, with an AttAUC of 1.0 indicating perfect alignment and 0.8 showing partial alignment. Darker red shades indicate higher attribution of text elements to image regions.}
    \label{fig:accuracy-evaluation}
\end{figure}
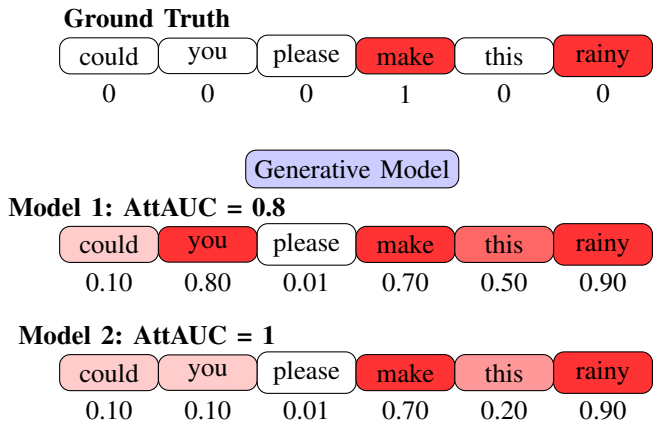

For example, Fig.~\ref{fig:accuracy-evaluation} shows that the ground truth identifies the words \emph{“make”} and \emph{“rainy”} as the most relevant text elements for generating an image. The metric \textbf{AttAUC} evaluates the model's alignment with the ground truth, where:
\begin{itemize}
    \item \textbf{AttAUC = 1.0:} Perfect alignment.
    \item \textbf{Lower AttAUC (0.8):} Less accurate~\cite{powers2020evaluation}.
\end{itemize}

The heatmaps produced by the model visually represent as you see in Fig.~\ref{fig:accuracy-evaluation}
these attention scores, where darker red shades indicate higher attribution to specific text elements. An accurate model should correctly assign higher attention to the text elements that most influence the visual output, improving the interpretability and reliability of the model~\cite{fong2017interpretable}.

\subsubsection{Stability}
Stability in an explainable model refers to its ability to provide consistent explanations even when minor changes are made to the input data~\cite{samek2017explainable}. This means that small perturbations in the input should not significantly affect the model’s predictions or explanations. Stability is a key property for fostering trust in explainable artificial intelligence (XAI) systems~\cite{doshi2017towards}. It reassures users that the model's interpretations are not arbitrary or overly sensitive to irrelevant variations~\cite{lipton2018mythos}.

\begin{figure}[H]
    \centering
    \begin{tikzpicture}
    \definecolor{whitebox}{rgb}{1, 1, 1}
    \definecolor{lightpink}{rgb}{1, 0.8, 0.8}
    \definecolor{mediumred}{rgb}{1, 0.6, 0.6}
    \definecolor{red}{rgb}{1, 0.4, 0.4}
    \definecolor{darkred}{rgb}{1, 0.2, 0.2}

    \node[align=center, font=\bfseries] at (-2, 4.5) {Attribution Stability Test};

    \node[draw, rounded corners, fill=whitebox, minimum width=1.3cm] (w1) at (-5.5, 4) {could};
    \node[draw, rounded corners, fill=whitebox, minimum width=1.3cm] (w2) at (-4.2, 4) {you};
    \node[draw, rounded corners, fill=whitebox, minimum width=1.3cm] (w3) at (-2.9, 4) {please};
    \node[draw, rounded corners, fill=whitebox, minimum width=1.3cm] (w4) at (-1.6, 4) {make};
    \node[draw, rounded corners, fill=whitebox, minimum width=1.3cm] (w5) at (-0.3, 4) {this};
    \node[draw, rounded corners, fill=whitebox, minimum width=1.3cm] (w6) at (1, 4) {rainy};

    \node[draw, rounded corners, fill=blue!20, minimum width=2.5cm, minimum height=0.5cm] (pix2pix1) at (-2, 3) {Generative Model};

    \node[draw, rounded corners, fill=lightpink, minimum width=1.3cm] (j1) at (-5.5, 2) {could};
    \node[draw, rounded corners, fill=lightpink, minimum width=1.3cm] (j2) at (-4.2, 2) {you};
    \node[draw, rounded corners, fill=whitebox, minimum width=1.3cm] (j3) at (-2.9, 2) {please};
    \node[draw, rounded corners, fill=darkred, minimum width=1.3cm] (j4) at (-1.6, 2) {make};
    \node[draw, rounded corners, fill=mediumred, minimum width=1.3cm] (j5) at (-0.3, 2) {this};
    \node[draw, rounded corners, fill=darkred, minimum width=1.3cm] (j6) at (1, 2) {rainy};

    \node[draw, rounded corners, fill=whitebox, minimum width=1.2cm] (t1) at (-5.5, -0.5) {could};
    \node[draw, rounded corners, fill=whitebox, minimum width=1.2cm] (t2) at (-4.3, -0.5) {you};
    \node[draw, rounded corners, fill=whitebox, minimum width=1.2cm] (t3) at (-3.1, -0.5) {please};
    \node[draw, rounded corners, fill=whitebox, minimum width=1.2cm] (t4) at (-1.9, -0.5) {make};
    \node[draw, rounded corners, fill=whitebox, minimum width=1.2cm] (t5) at (-0.7, -0.5) {this};
    \node[draw, rounded corners, fill=whitebox, minimum width=1.2cm] (t6) at (0.5, -0.5) {rainy};
    \node[draw, rounded corners, fill=whitebox, minimum width=1.2cm] (t7) at (1.7, -0.5) {\#\#\#};

    \node[draw, rounded corners, fill=blue!20, minimum width=2.5cm, minimum height=0.5cm] (pix2pix) at (-1.5, -1.5) {Generative Model};

    \node[draw, rounded corners, fill=lightpink, minimum width=1.2cm] (b1) at (-5.5, -2.5) {could};
    \node[draw, rounded corners, fill=lightpink, minimum width=1.2cm] (b2) at (-4.3, -2.5) {you};
    \node[draw, rounded corners, fill=whitebox, minimum width=1.2cm] (b3) at (-3.1, -2.5) {please};
    \node[draw, rounded corners, fill=darkred, minimum width=1.2cm] (b4) at (-1.9, -2.5) {make};
    \node[draw, rounded corners, fill=mediumred, minimum width=1.2cm] (b5) at (-0.7, -2.5) {this};
    \node[draw, rounded corners, fill=darkred, minimum width=1.2cm] (b6) at (0.5, -2.5) {rainy};
    \node[draw, rounded corners, fill=lightpink, minimum width=1.2cm] (b7) at (1.7, -2.5) {\#\#\#};

    \node at (-5.5, 1.5) {0.10};
    \node at (-4.2, 1.5) {0.10};
    \node at (-2.9, 1.5) {0.01};
    \node at (-1.6, 1.5) {0.70};
    \node at (-0.3, 1.5) {0.20};
    \node at (1, 1.5) {0.90};

    \node at (-5.5, -3) {0.10};
    \node at (-4.2, -3) {0.10};
    \node at (-3, -3) {0.01};
    \node at (-2, -3) {0.70};
    \node at (-0.8, -3) {0.20};
    \node at (0.6, -3) {0.90};
    \node at (1.6, -3) {0.10};

    \draw[-, dashed, red] (w1) -- (pix2pix1);
    \draw[-, dashed, red] (w2) -- (pix2pix1);
    \draw[-, dashed, red] (w3) -- (pix2pix1);
    \draw[-, dashed, red] (w4) -- (pix2pix1);
    \draw[-, dashed, red] (w5) -- (pix2pix1);
    \draw[-, dashed, red] (w6) -- (pix2pix1);

    \draw[->, dashed, red] (pix2pix1) -- (j1);
    \draw[->, dashed, red] (pix2pix1) -- (j2);
    \draw[->, dashed, red] (pix2pix1) -- (j3);
    \draw[->, dashed, red] (pix2pix1) -- (j4);
    \draw[->, dashed, red] (pix2pix1) -- (j5);
    \draw[->, dashed, red] (pix2pix1) -- (j6);

    \draw[-, dashed, red] (t1) -- (pix2pix);
    \draw[-, dashed, red] (t2) -- (pix2pix);
    \draw[-, dashed, red] (t3) -- (pix2pix);
    \draw[-, dashed, red] (t4) -- (pix2pix);
    \draw[-, dashed, red] (t5) -- (pix2pix);
    \draw[-, dashed, red] (t6) -- (pix2pix);
    \draw[-, dashed, red] (t7) -- (pix2pix);

    \draw[->, dashed, red] (pix2pix) -- (b1);
    \draw[->, dashed, red] (pix2pix) -- (b2);
    \draw[->, dashed, red] (pix2pix) -- (b3);
    \draw[->, dashed, red] (pix2pix) -- (b4);
    \draw[->, dashed, red] (pix2pix) -- (b5);
    \draw[->, dashed, red] (pix2pix) -- (b6);
    \draw[->, dashed, red] (pix2pix) -- (b7);

    \end{tikzpicture}
    \caption{Stability evaluation with slightly different text inputs. The model is provided with two text prompts: one with ``could you please make this rainy weather.'' and another with ``could you please make this rainy.'' The resulting attention heatmaps show similar focus on the words ``make'' and ``rainy,'' demonstrating the stability of the model’s output.}
    \label{fig:stability-evaluation}
\end{figure}
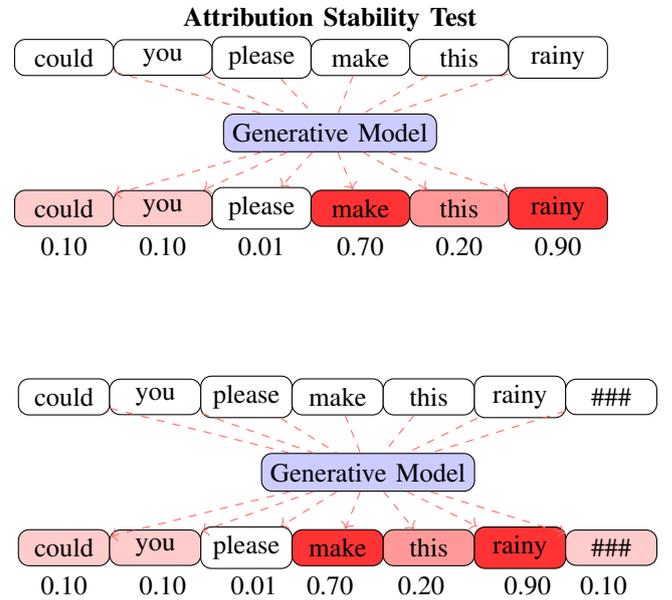

In the example image, the attention scores are visualized for two slightly different text inputs. The heatmaps show how the model assigns attention to each word. A stable model, as demonstrated in Fig.~\ref{fig:stability-evaluation}, assigns similar importance to the words \textit{make} and \textit{rainy} in both inputs, even though the sign \textit{\#\#\#} was added in one case. This consistency is critical for ensuring that the model behaves predictably across minor input variations~\cite{ahmadi2024explainability}. Without such stability, the explanations generated by the model could lead to confusion or misinterpretation, especially in high-stakes scenarios such as medical diagnosis or financial decision-making~\cite{ribeiro2016should}.

To quantify stability, we use the \textbf{Jaccard index}, which measures the similarity between two sets of model predictions. For two sets \( A \) and \( B \), the Jaccard index is calculated as follows in Eq.~\ref{eq:jaccard}:

\begin{equation}
\label{eq:jaccard}
\text{Jaccard}(A, B) = \frac{|A \cap B|}{|A \cup B|}
\end{equation}

This metric reflects the proportion of shared elements between the sets, where values closer to 1 indicate greater stability and similarity in explanations~\cite{samek2017explainable}. A higher Jaccard index confirms the model's ability to yield stable outputs, thus providing consistent and reliable interpretations~\cite{vaswani2017attention}. By ensuring stability, we not only enhance the robustness of the model but also its usability and credibility in real-world applications~\cite{doshi2017towards}.

\subsubsection{Consistency}

\begin{figure}[H]
    \centering
    \begin{tikzpicture}
    \definecolor{whitebox}{rgb}{1, 1, 1}
    \definecolor{lightpink}{rgb}{1, 0.8, 0.8}
    \definecolor{mediumred}{rgb}{1, 0.6, 0.6}
    \definecolor{red}{rgb}{1, 0.4, 0.4}
    \definecolor{darkred}{rgb}{1, 0.2, 0.2}

    \node[align=center, font=\bfseries] at (-4., 6) {Attribution Consistency Test};

    \node[draw, rounded corners, fill=whitebox, minimum width=1.3cm] (b1) at (-5.5, 5.5) {could};
    \node[draw, rounded corners, fill=whitebox, minimum width=1.3cm] (b2) at (-4.2, 5.5) {you};
    \node[draw, rounded corners, fill=whitebox, minimum width=1.3cm] (b3) at (-2.9, 5.5) {please};
    \node[draw, rounded corners, fill=whitebox, minimum width=1.3cm] (b4) at (-1.6, 5.5) {make};
    \node[draw, rounded corners, fill=whitebox, minimum width=1.3cm] (b5) at (-0.3, 5.5) {this};
    \node[draw, rounded corners, fill=whitebox, minimum width=1.3cm] (b6) at (1, 5.5) {rainy};

    \node[draw, rounded corners, fill=blue!20, minimum width=1.5cm, minimum height=0.5cm] (pix2pix1) at (-5, 4.2) {Generative Model(1)};

    \node[draw, rounded corners, fill=lightpink, minimum width=1.3cm] (t1) at (-5.5, 3.5) {could};
    \node[draw, rounded corners, fill=lightpink, minimum width=1.3cm] (t2) at (-4.2, 3.5) {you};
    \node[draw, rounded corners, fill=whitebox, minimum width=1.3cm] (t3) at (-2.9, 3.5) {please};
    \node[draw, rounded corners, fill=darkred, minimum width=1.3cm] (t4) at (-1.6, 3.5) {make};
    \node[draw, rounded corners, fill=mediumred, minimum width=1.3cm] (t5) at (-0.3, 3.5) {this};
    \node[draw, rounded corners, fill=darkred, minimum width=1.3cm] (t6) at (1, 3.5) {rainy};

    \node[draw, rounded corners, fill=blue!20, minimum width=1.5cm, minimum height=0.5cm] (pix2pix2) at (-5, 2.2) {Generative Model(2)};

    \node[draw, rounded corners, fill=lightpink, minimum width=1.3cm] (w1) at (-5.5, 1.5) {could};
    \node[draw, rounded corners, fill=darkred, minimum width=1.3cm] (w2) at (-4.2, 1.5) {you};
    \node[draw, rounded corners, fill=whitebox, minimum width=1.3cm] (w3) at (-2.9, 1.5) {please};
    \node[draw, rounded corners, fill=darkred, minimum width=1.3cm] (w4) at (-1.6, 1.5) {make};
    \node[draw, rounded corners, fill=mediumred, minimum width=1.3cm] (w5) at (-0.3, 1.5) {this};
    \node[draw, rounded corners, fill=darkred, minimum width=1.3cm] (w6) at (1, 1.5) {rainy};

    \node[draw, rounded corners, fill=blue!20, minimum width=1.5cm, minimum height=0.5cm] (pix2pix3) at (-5, 0.2) {Generative Model(3)};

    \node[draw, rounded corners, fill=lightpink, minimum width=1.3cm] (j1) at (-5.5, -0.5) {could};
    \node[draw, rounded corners, fill=lightpink, minimum width=1.3cm] (j2) at (-4.2, -0.5) {you};
    \node[draw, rounded corners, fill=whitebox, minimum width=1.3cm] (j3) at (-2.9, -0.5) {please};
    \node[draw, rounded corners, fill=darkred, minimum width=1.3cm] (j4) at (-1.6, -0.5) {make};
    \node[draw, rounded corners, fill=mediumred, minimum width=1.3cm] (j5) at (-0.3, -0.5) {this};
    \node[draw, rounded corners, fill=darkred, minimum width=1.3cm] (j6) at (1, -0.5) {rainy};

    \node at (-5.5, 3) {0.10};
    \node at (-4.2, 3) {0.10};
    \node at (-2.9, 3) {0.01};
    \node at (-1.6, 3) {0.70};
    \node at (-0.3, 3) {0.20};
    \node at (1, 3) {0.90};

    \node at (-5.5, 1) {0.10};
    \node at (-4.2, 1) {0.10};
    \node at (-2.9, 1) {0.01};
    \node at (-1.6, 1) {0.70};
    \node at (-0.3, 1) {0.20};
    \node at (1, 1) {0.90};

    \node at (-5.5, -1) {0.10};
    \node at (-4.2, -1) {0.10};
    \node at (-2.9, -1) {0.01};
    \node at (-1.6, -1) {0.70};
    \node at (-0.3, -1) {0.20};
    \node at (1, -1) {0.90};

    \end{tikzpicture}
    \caption{Consistency analysis of the Instruct-Pix2Pix model in response to the prompt ``Could you please make this rainy.'' The heatmaps show the influence of each word on the model's output across three iterations.}
    \label{fig:pix2pix-consistency}
\end{figure}
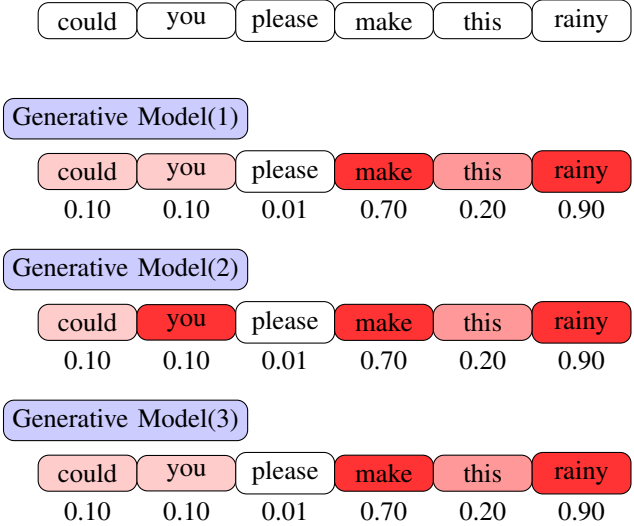

In evaluating the consistency of our model, we assess whether it produces stable and reliable outputs when the same input is provided multiple times. Consistency ensures that the model behaves predictably, reinforcing trust in its outputs and usability across diverse scenarios~\cite{ribeiro2016should}.
As shown in Fig.~\ref{fig:pix2pix-consistency}, when the input phrase ``Could you please make this rainy?'' was fed into the model multiple times, the output was stable across iterations. This consistency is crucial for applications where repeatable and reliable results are required. In our case, the model consistently assigns higher weights to the words ``make'' and ``rainy'', demonstrating that it understands the importance of these words in generating the desired output~\cite{vaswani2017attention}. Such behavior underscores the model's ability to focus on the semantic core of the input, a trait essential for maintaining coherence in practical use cases~\cite{lipton2018mythos}.
In practical applications, such as automated content generation, weather simulation, or interactive systems, users depend on models to produce results that do not fluctuate unexpectedly. Any inconsistencies could undermine user trust and compromise the system's overall effectiveness. By demonstrating consistent outputs, our model not only enhances user confidence but also establishes a foundation for further optimization and integration into more complex systems. This ability to maintain consistency across iterations directly contributes to the robustness and scalability of the model, making it well-suited for deployment in dynamic real-world environments~\cite{doshi2017towards}.

\subsubsection{Fidelity}

Fidelity is a metric used to evaluate how well an explainable model aligns with a black-box model. The predictions of the black-box model, denoted as \( f(X_i) \), and the explainable model, \( g(X_i) \), for the \(i\)-th data perturbation, are compared. Various loss functions and coefficient metrics quantify the similarity between the models' predictions, measuring how effectively the explainable model approximates the black-box model.

The \textbf{coefficient of determination} \( R^2 \) is an accuracy-based metric that measures how well the explainable model correlates with the black-box model~\cite{draper1998applied} and is defined as Eq.~\ref{eq:coefficient_of_determination}:

\begin{equation} \label{eq:coefficient_of_determination}
R^2 = 1 - \frac{\sum_{i=1}^{N_p} (f(X_i) - g(X_i))^2}{\sum_{i=1}^{N_p} (f(X_i) - \bar{f}(X_i))^2}
\end{equation}

where \( \bar{f}(X_i) \) represents the average of \( f(X_i) \). \( R^2 \) values close to 1 indicate a strong alignment between the two models.

For scenarios where weighting is necessary, \( R_w^2 \), the \textbf{weighted coefficient of determination}, provides an alternative version of \( R^2 \)~\cite{ahmadi2024explainability, biometrics19763211hocking1976analysis} defined as Eq.~\ref{eq:weighted_coefficient_of_determination}:

\begin{equation} \label{eq:weighted_coefficient_of_determination}
R_w^2 = 1 - \frac{\sum_{i=1}^{N_p} (f(X_i) - g(X_i))^2}{\sum_{i=1}^{N_p} (f(X_i) - \bar{f_w}(X_i))^2}
\end{equation}

where \( \bar{f_w}(X_i) \) represents the weighted average of \( f(X_i) \). These accuracy-based metrics, \( R^2 \) and \( R_w^2 \), provide insights into how well the explainable model approximates the black-box model.

To compensate for both sample size and the number of variables, the \textbf{weighted adjusted coefficient of determination} \( \hat{R}_w^2 \)~\cite{ahmadi2024explainability} is used, as Eq.~\ref{eq:weighted_adjusted_coefficient_of_determination}:

\begin{equation} \label{eq:weighted_adjusted_coefficient_of_determination}
\hat{R}_w^2 = 1 - (1 - R_w^2) \left[ \frac{N_p - 1}{N_p - N_s - 1} \right]
\end{equation}

To measure the error or risk between the models' predictions, error-based metrics, including Weighted Mean Squared Error (WMSE) and Weighted Mean Absolute Error (WMAE), are used~\cite{willmott2005advantages}. These metrics quantify the weighted difference between the predictions of the explainable model and those of the black-box model. Lower values indicate better alignment. The definitions of WMSE and WMAE are as Eqs.~\ref{eq:mse} and~\ref{eq:mae}:

\begin{equation} \label{eq:mse}
\text{WMSE} = \frac{\sum_{i=1}^{N_p} w_i \cdot (f(X_i) - g(X_i))^2}{\sum_{i=1}^{N_p} w_i}
\end{equation}

\begin{equation} \label{eq:mae}
\text{WMAE} = \frac{\sum_{i=1}^{N_p} w_i \cdot |f(X_i) - g(X_i)|}{\sum_{i=1}^{N_p} w_i}
\end{equation}

Here, \( w_i \) denotes the weight for each data point \( i \), \( f(X_i) \) represents the predictions from the explainable model, \( g(X_i) \) represents the predictions from the black-box model, and \( N_p \) is the total number of data points.

Additionally, the \textbf{mean \( L_1 \)} and \textbf{mean \( L_2 \)} losses, defined below, are commonly used to assess prediction discrepancies~\cite{goodfellow2016deep, ahmadi2024explainability} as Eqs.~\ref{eq:l1_loss} and~\ref{eq:l2_loss}:

\begin{equation} \label{eq:l1_loss}
L_1 = \frac{1}{N_p} \sum_{i=1}^{N_p} |f(X_i) - g(X_i)|
\end{equation}

\begin{equation} \label{eq:l2_loss}
L_2 = \frac{1}{N_p} \sum_{i=1}^{N_p} (f(X_i) - g(X_i))^2
\end{equation}

These error-based metrics, including MSE, MAE, \( L_1 \), and \( L_2 \), quantify the differences between the predicted scores of the black-box model and the explainable model, reflecting the risk of prediction discrepancies.

We compute fidelity across various scenarios, as summarized in the following tables. Fidelity is measured by comparing the explainable model's predictions to those of the black-box model. We examine how different perturbations in the input text affect fidelity scores, as well as how various diffusion models and distance metrics for text-versus-text and image-versus-image comparisons perform when using linear regression and Bayesian ridge as surrogate models.

\section{Experimental Results}

This section evaluates the proposed ability to enhance explainability in instruction-based image editing. We analyze how the structure of textual prompts affects interpretability, how input images influence the clarity of explanations, and how the proposed solution performs across diverse scenarios. Experiments were conducted using a range of datasets and three state-of-the-art diffusion models—Instruct-Pix2Pix~\cite{an2023fine}, Img2Img-Turbo~\cite{parmar2024one}, and Diffusers-Inpaint~\cite{rombach2022high}—to ensure comprehensive assessment.

\subsection{Qualitative Results}

The qualitative evaluation demonstrates the interpretability of the proposed framework using scenario-based testing and visualizations. Leveraging the Operational Design Domain (ODD)~\cite{zhang2024odd}, we generated diverse testing scenarios encompassing environmental factors like weather transformations (e.g., rain, fog, snow) and non-environmental variables (e.g., object attributes, human-centric factors, temporal conditions). 

Heatmaps illustrate how specific keywords influence image edits, as shown in Tables~\ref{tab:weather_transformations} and~\ref{tab:table_of_co_2}. Further, we analyzed input prompt explainability by mapping word contributions to visual changes, highlighting the weight of keywords such as ``snowing'' in the editing process. Using t-SNE scatter plots, we qualitatively examined the clustering of images generated from perturbed prompts, revealing the impact of word-level variations on the generated outputs (Figs.~\ref{fig:TSNE1}).

\subsubsection{Input Prompt Explainability}

The proposed scenario-based testing framework leverages the system's Operational Design Domain (ODD) to generate scenarios that comprehensively test Automated Driving Systems (ADS) under a wide array of conditions~\cite{koopman2016challenges, zhang2024odd}. 

The concept of ODD, as highlighted in recent high-risk AI frameworks, is fundamental to understanding and ensuring the operational boundaries and limitations of ADS~\cite{wang2024survey}. As noted in broader discussions on high-risk AI systems, the ODD provides a structured way to classify environmental and situational variables, helping to consistently assess and validate system performance across various conditions~\cite{stettinger2024trustworthiness}.

Tables~\ref{tab:weather_transformations} and~\ref{tab:table_of_co_2} are developed based on the Operational Design Domain (ODD) framework to systematically simulate and assess various range of environmental and situational conditions affecting the performance of Automated Driving Systems (ADS). 

The ODD, central to high-risk AI frameworks, defines ADS's operational boundaries and limitations, enabling the structured creation of diverse testing scenarios~\cite{koopman2016challenges}. Table~\ref{tab:weather_transformations} is generated using a variety of instructions, focusing on environmental conditions and classifying scenarios by weather states such as rain, fog, and snow, ensuring thorough evaluation under diverse environmental factors~\cite{zhang2024odd}. 

Table~\ref{tab:table_of_co_2} incorporates various instructions to address non-environmental variables, including person-related attributes (e.g., age, gender, and skin color), object variations (e.g., vehicle colors and traffic light configurations), and temporal factors (e.g., day and night scenarios), ensuring thorough testing across a range of human-centric and object-driven conditions~\cite{stettinger2024trustworthiness}. 

We implemented interpretability techniques using various image editing models, including Instruct-Pix2Pix~\cite{an2023fine}, Img2Img-Turbo~\cite{parmar2024one} and Diffusers-Inpaint~\cite{rombach2022high}. We based our approach on the analysis provided in the tables and incorporated heatmaps to demonstrate how specific keywords influence image modifications~\cite{koopman2016challenges}. This structured methodology is aligned with high-risk AI frameworks, such as the EU AI Act, which emphasizes trustworthiness, safety, and regulatory compliance throughout the system's lifecycle~\cite{outeda2024eu}.

Additionally, it strengthens the robustness, reliability, and transparency of the Automated Decision System (ADS) evaluation process. Our design aims to enhance the trustworthiness of high-risk AI systems through instruction-based image editing models.

\subsubsection{Input Image Explainability}

In this part, we explore the interpretability of image editing models by analyzing the weight that specific words in a prompt exert on edited images. Using a fixed prompt, we applied edits to 10 images and utilized SMILE (Statistical Model-agnostic Interpretability with Local Explanations) to generate heatmaps for each edited image. These heatmaps visualize the weight of each word in the prompt, showing which parts of the images are most influenced by specific terms during the editing process.This approach allows us to identify and quantify the relationship between textual instructions and corresponding visual modifications, highlighting the importance of semantic alignment in text-to-image models.

By aggregating the weights from the heatmaps across all ten images, we created a box plot to display the influence distribution for each word in the prompt. 
This method evaluates how specific words are crucial in editing image models that providing an understanding of how text prompts affect visual modification.

\begin{figure}[H]
    \centering
    \includegraphics[width=1\linewidth]{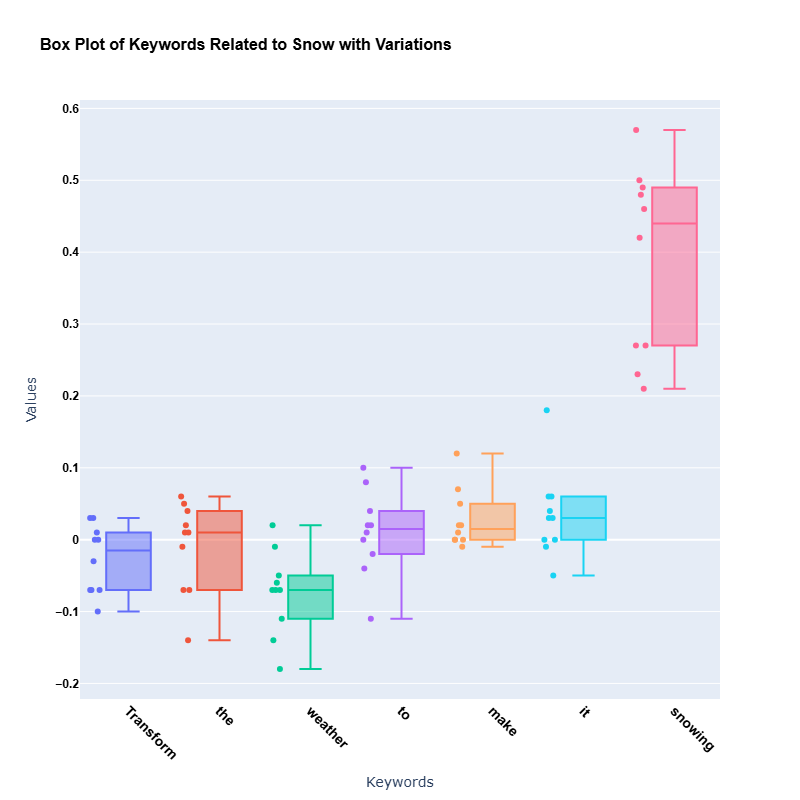}
    \caption{The box plot illustrates the distribution of word weights for each term in the prompt ``Transform the weather to make it snowing'' across 10 edited images with 30 perturbations. Key terms like ``snowing'' have higher median weights and variability, indicating their strong influence in the editing process compared to other words.}
    \label{fig:snow_keywords_boxplot}
\end{figure}

This analysis provides valuable insights into the interpretative mechanics of the model, showing how individual words shape the final edited output. As shown in Fig.~\ref{fig:snow_keywords_boxplot}, the box plot reveals the range and concentration of influence each word exerts.
The keyword ``snowing'' has the most substantial influence compared to other words in the prompt, indicating a significant effect in the model's editing process for creating snow-related adjustments.

\subsubsection{t-SNE Scatter Plot of Image Embeddings}

In this section, we present the qualitative results of our model using perturbed prompts. To better understand the influence of various prompt structures on the generated images, we visualized the results using a 2D plot. This plot displays the distances between images generated from different perturbations of the same base scenario. Each point in the plot represents an image, with the red dot corresponding to the original image and the blue points representing the perturbed images.

We systematically modified the prompt by adding or removing keywords, such as ``snowing''. As shown in the plot, perturbations that include the keyword ``snowing'' lead to significant changes in the generated image, transforming the scene into snowy. In contrast, when the perturbation does not include this keyword, the resulting images do not depict snow.

\begin{figure*}
    \centering
    \includegraphics[width=1\linewidth]{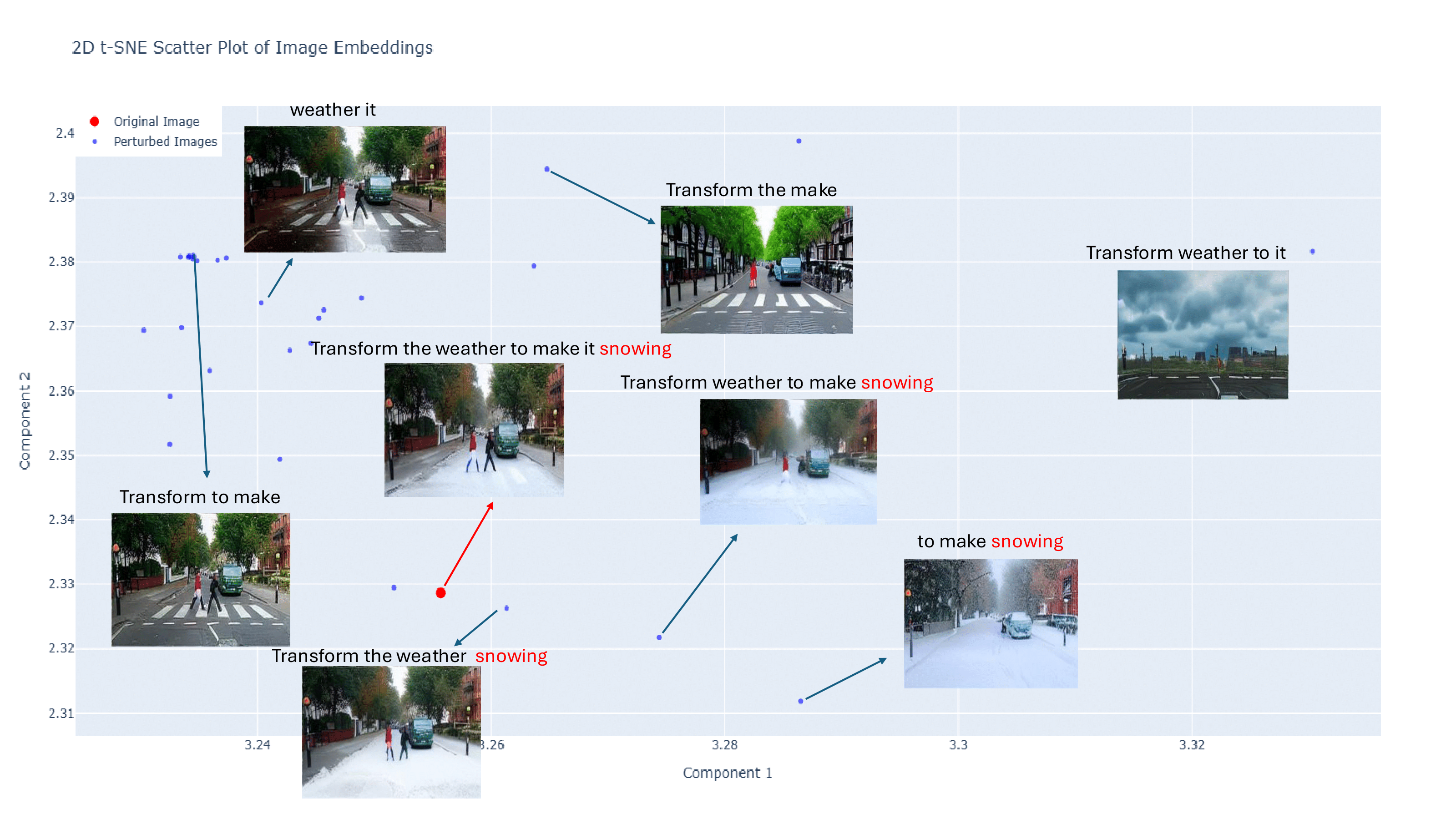}
    \caption{2D visualization of the qualitative results from perturbed prompts, showing the impact of including or omitting the keyword ``snowing'' on the generated images. The red dot represents the original image, while the blue points represent perturbed images. Perturbations that include the keyword ``snowing'' result in snowy scenes, while those without the keyword remain closer to the original non-snowy scene.}
    \label{fig:TSNE1}
\end{figure*}

The 2D plot shown in Fig.~\ref{fig:TSNE1} demonstrates how close or distant the images are based on their visual features, with perturbations containing the keyword ``snowing'' forming clusters in the snowy regions. In contrast, others remain closer to the original, non-snowy scene. This visualization allows us to qualitatively assess the impact of each word in the prompt and how the model responds to different perturbations.

\subsection{Quantitative Results}

In this part, we apply some evaluation metrics, including accuracy, stability, consistency, fidelity, and computation complexity, to asses our explainability method across different image editing models and different scenarios.

\subsubsection{Accuracy}

To evaluate the performance of instruction-based image editing models, we tested ten different prompts with a single image and ten different images with a single instruction by 30 perturbations across. Heatmaps were extracted to display the weights of each keyword in the instructions across various models, including Instruct-Pix2Pix, Img2Img-Turbo and Diffusers-inpaint. A ground truth was defined, assigning weights of 1 to keywords in the instructions and 0 to other words. These ground truth values were compared with the extracted heatmaps to assess model performance.

To quantify accuracy, we employed multiple metrics, including Attention Accuracy (ATT ACC), F1-Score for Attention (ATT F1), and Area Under the Receiver Operating Characteristic Curve for Attention (ATT AUROC). The results were averaged across the ten different prompts and images, and each model's outcomes are summarized in Table~\ref{tab:AccuraCy_models}, showcasing their performance based on these metrics.

\begin{table}[H]
\centering
\caption{Performance metrics (ATT ACC, ATT F1, and ATT AUROC) for various\\models, evaluated under two conditions: different prompts and different images.}
\label{tab:AccuraCy_models}
\renewcommand{\arraystretch}{1.3}
\scalebox{0.8}{
\begin{tabular}{ccccccc}
\hline \hline
\textbf{Model name} & \textbf{(ATT ACC)} & \textbf{(ATT F1)} & \textbf{(ATT AUROC)} \\ \hline
I-Pix2Pix (different prompts) & 0.7845 & 0.6869 & 0.8943 \\ 
I-Pix2Pix (different images) & 0.9571 & 0.9166 & 1.000 \\ 
Diffusers\_I (different prompts) & 0.6621 & 0.1167 & 0.4417 \\
Diffusers\_I (different images) & 0.7714 & 0.1000 & 0.4000 \\
I2I-Turbo (different prompts) & 0.7944 & 0.6824 & 0.8400 \\  
I2I-Turbo (different images) & 0.8286 & 0.6333 & 1.000 \\ 
\hline \hline
\end{tabular}
}
\end{table}

\subsubsection{Stability}

To quantify stability, we use the Jaccard index, a metric that measures the similarity between sets by comparing their overlap. We modify an image with ten different prompts for image editing models such as Instruct-Pix2Pix, Img2Img-Turbo and Diffusers-inpaint, extracting the coefficient and weight of each word. Then, by adding `\#\#\#` at the end of each prompt as input text, we extract the coefficients again and compute the Jaccard index to compare them. Finally, we calculate the average metric for each model and present the results in Table~\ref{tab:stability_models}.

\begin{table}[H]
\centering
\caption{Stability across different models for ten prompts and 30 perturbations}
\label{tab:stability_models}
\renewcommand{\arraystretch}{1.3}
\scalebox{0.8}{
\begin{tabular}{ccc}
\hline \hline
\textbf{Model name}  & \textbf{Jaccard Index}  \\ \hline
I-Pix2Pix & 0.85 & \\
Diffusers\_I & 0.85 & \\
I2I-Turbo & 0.85 & \\ \hline \hline
\end{tabular}
}
\end{table}

\subsubsection{Consistency}

For compute consistency, we ran the same code with 30 perturbations for each one over 1000 iterations, ensuring that the outputs remained consistent across these repetitions for image editing models such as Instruct-Pix2Pix, Img2Img-Turbo and Diffusers-inpaint, as shown in table~\ref{tab:consistency_models}. We computed variance and standard deviation metrics for each word coefficient. This step is essential to confirm that the model’s predictions are not random or heavily dependent on initialization factors but rather deterministic and robust.

\begin{table}[H]
\centering
\caption{Consistency metrics for different diffusion models on the prompt\\``Transform the weather to make it snowing.''}
\label{tab:consistency_models}
\renewcommand{\arraystretch}{1.3}
\scalebox{0.8}{
\begin{tabular}{ccc}
\hline \hline
\textbf{Model name}  & \textbf{Variance} & \textbf{Standard Deviation} \\ \hline
I-Pix2Pix & 0.0161 & 0.1271 \\
Diffusers\_I & 0.0776 & 0.0060 \\
I2I-Turbo & 0.0001 & 0.0081 \\ \hline \hline
\end{tabular}
}
\end{table}

\subsubsection{Fidelity for Different Instruct image editing diffusion based Models (IED)}

The fidelity computation for various IED models applied to the text prompt ``Transform the weather to make it snowing'' is displayed below. We compare several models, including Instruct-Pix2Pix, Img2Img-Turbo, and Diffusers-inpaint, using 64 perturbations. The comparisons are made using a weighted linear regression as the surrogate model and the Wasserstein distance to compute distances across metrics such as MSE, (R$^2_\omega$), MAE, mean \(L_1\), and mean \(L_2\) losses. The results are presented in Table.~\ref{tab:fidelity_models}.

\begin{table}[H]
\centering
\caption{Fidelity metrics for different diffusion models on the prompt\\``Transform the weather to make it snowing.''}
\label{tab:fidelity_models}
\renewcommand{\arraystretch}{1.3}
\scalebox{0.8}{
\begin{tabular}{ccccccc}
\hline \hline
\textbf{Model name} & \textbf{(MSE)} & \textbf{(R$^2_\omega$)} & \textbf{(MAE)} & \textbf{L1} & \textbf{L2} & \textbf{(R$^2_{\hat{\omega}}$)} \\ \hline
I-Pix2Pix & 0.0120 & 0.7208 & 0.0812 & 0.0790 & 0.0124 & 0.6859 \\
Diffusers\_I & 0.0317 & 0.1255 & 0.1359 & 0.1529 & 0.0410 & 0.0161  \\
I2I-Turbo & 0.0193 & 0.6225 & 0.1078 & 0.1076 & 0.0187 & 0.5753 \\
\hline \hline
\end{tabular}
}
\end{table}

As shown in the table~\ref{tab:fidelity_models}, the fidelity for the Diffusers-inpaint model was calculated, but its value is relatively low due to the model's poor performance. For this reason, fidelity was not calculated for the other scenarios involving this model.

\subsubsection{Fidelity Across Different number of Text Perturbations}

Tables.~\ref{tab:fidelity_per}, ~\ref{tab:fidelity_per_turbo} shows the Results of fidelity computation for different numbers of perturbations using the Instruct-Pix2Pix and Img2Img-Turbo methods. The comparisons are made using a weighted linear regression as the surrogate model and the Wasserstein distance to compute distances, using metrics such as mean squared error (MSE), coefficient of determination (R$^2_\omega$), mean absolute error (MAE),  loss metrics of mean (\(L_1\) and mean \(L_2\)):

\begin{table}[H]
\centering
\caption{Performance metrics for different numbers of perturbations in Instruct-Pix2Pix}
\label{tab:fidelity_per}
\renewcommand{\arraystretch}{1.3}
\scalebox{0.8}{
\begin{tabular}{rcccccc}
\hline \hline
\textbf{\#Perturb} & \textbf{(MSE)} & \textbf{(R$^2_\omega$)} & \textbf{(MAE)} & \textbf{L1} & \textbf{L2} & \textbf{(R$^2_{\hat{\omega}}$)} \\ \hline
32  & 0.0178 & 0.7770 & 0.1040 & 0.1004 & 0.0161 & 0.7119 \\ 
64  & 0.0120 & 0.7208 & 0.0812 & 0.0790 & 0.0124 & 0.6859 \\ 
128  &0.0220  &0.6816  &0.1129  &0.1090  &0.0207  &0.6631  \\ 
256  & 0.0080 & 0.5110 & 0.0512 & 0.0509 & 0.0084 & 0.4972 \\ 
\hline \hline
\end{tabular}
}
\end{table}

\begin{table}[H]
\centering
\caption{Performance metrics for different numbers of perturbations in I2I-Turbo}
\label{tab:fidelity_per_turbo}
\renewcommand{\arraystretch}{1.3}
\scalebox{0.8}{
\begin{tabular}{rcccccc}
\hline \hline\textbf{\#Perturb} & \textbf{(MSE)} & \textbf{(R$^2_\omega$)} & \textbf{(MAE)} & \textbf{L1} & \textbf{L2} & \textbf{(R$^2_{\hat{\omega}}$)} \\ \hline
32  & 0.0421 & 0.5273 & 0.1635 & 0.1484 & 0.0369 & 0.3894 \\ 
64  & 0.0193 & 0.6225 & 0.1078 & 0.1076 & 0.0187 & 0.5753 \\ 
128  &0.0316  &0.3879  &0.1412  &0.1330  &0.0293  &0.3522  \\ 
256  & 0.0248 & 0.6025 & 0.1275 & 0.1375 & 0.0305 & 0.5912 \\ 
\hline \hline
\end{tabular}
}
\end{table}

\subsubsection{Fidelity for Different Distance Metrics and Surrogate Models}

Fidelity is also computed by comparing various distance metrics for the Instruct-Pix2Pix model with text-versus-text and image-versus-image comparisons. The surrogate models used are Weighted Linear Regression and Bayesian Ridge (BayLIME), as shown in Table.~\ref{tab:fidelity_dist}. We explore different combinations of distance metrics such as Cosine, Wasserstein Distance (WD). Fidelity is measured using MSE, (R$^2_\omega$), MAE, mean \(L_1\), and mean \(L_2\) losses.

\begin{table}[H]
\centering
\caption{Fidelity results for different distance measures with 30 perturbations}
\label{tab:fidelity_dist}
\renewcommand{\arraystretch}{1.2}
\scalebox{0.75}{ 
    \begin{tabular}{llcccccc} 
    \hline \hline
    \multicolumn{2}{c}{\textbf{WLR}} & \multicolumn{6}{c}{\textbf{Fidelity Metrics}} \\
    \cmidrule(lr){1-2} \cmidrule(lr){3-8}
    \textbf{T vs T} & \textbf{I vs I} &  \textbf{(MSE)} & \textbf{(R$^2_\omega$)} & \textbf{(MAE)} & \textbf{L1} & \textbf{L2} & \textbf{(R$^2_{\hat{\omega}}$)} \\
    \midrule
    Cosine & Cosine & 0.0494 & 0.7626 & 0.1555 & 0.2865 & 0.2309 & 0.6167 \\
    Cosine & WD & 0.0022 & 0.8557 & 0.0215 & 0.2530 & 0.1148 & 0.7980 \\
    WD & WD & 0.0128 & \textbf{0.9495} & 0.0674 & 0.2865 & 0.5243 & \textbf{0.9292} \\
    WD & Cosine & \textbf{0.0022} & 0.8558 & \textbf{0.0216} & 0.2530 & 0.1149 & 0.8558 \\
    WD+C & WD+C & 0.0411 & 0.8744 & 0.1427 & 0.8484 & 1.5264 & 0.8241 \\
    \midrule
    \multicolumn{2}{c}{\textbf{BayLIME}} & \multicolumn{6}{c}{} \\
    \midrule
    Cosine & Cosine & 0.0524 & 0.7097 & 0.1862 & 0.2607 & 0.1363 & 0.5936 \\
    Cosine & WD & 0.2314 & 0.6064 & 0.3942 & 0.5239 & 0.4739 & 0.6064 \\
    WD & WD & 0.0148 & \textbf{0.9485} & 0.0674 & 0.8065 & 0.8052 & \textbf{0.9278} \\
    WD & Cosine & \textbf{0.0001} & 0.8369 & \textbf{0.0049} & \textbf{0.0501} & \textbf{0.0044} & 0.7715 \\
    WD+C & WD+C & 0.0421 & 0.8711 & 0.1139 & 0.8282 & 1.4246 & 0.8195 \\
    \hline \hline
    \end{tabular} 
} 
\end{table} 

\subsubsection{Computation complexity}

We apply our model to three types of instruction-based image editing frameworks: Instruct-pix2pix, Img2Img-Turbo, and Diffusers-inpaint and measure the time consumption across different explainability methods, including LIME, SMILE and Bay-LIME, with 60 perturbations performed on the same computer to ensure consistent hardware conditions.

The models show apparent differences in their computational demands, mainly because of how they calculate similarity and assign weights. SMILE, in particular, uses the Wasserstein distance to calculate text distances, making it much more computationally expensive than LIME and Bay-LIME. As shown in Table~\ref{tab:execution_times}, the execution times for SMILE are consistently higher across all three frameworks. Similar trends are observed in the other frameworks, where SMILE's computational overhead is evident.

Wasserstein distance is excellent for capturing detailed differences between distributions, especially in text embeddings, but it comes at a cost \cite{peyre2019computational}. For high-dimensional text data, computing the Wasserstein-2 distance involves working with covariance matrices, which requires time proportional to $O(d^3)$, where $d$ is the embedding size. When multiple samples ($N$) are considered, the computational complexity quickly scales to $O(Nd^3)$. In contrast, LIME uses more straightforward similarity measures like cosine similarity, which scale much more efficiently at $O(Nd)$.

To further elaborate, the complexity difference arises because Wasserstein-2 distance calculations involve operations on covariance matrices for $d$-dimensional Gaussians~\cite{cuturi2014fast}. While this higher computational effort can yield more robust and stable explanations, as seen in SMILE, it significantly increases execution time. On the other hand, LIME's reliance on cosine similarity is computationally lighter and faster, as it avoids such matrix operations.

SMILE requires more computational resources and is inherently more complex, but it delivers highly detailed and stable explanations. This makes it particularly well-suited for tasks where interpretability and precision are critical despite its extra computational costs.

\begin{table}[H]
\centering
\caption{Execution times (in seconds) for each framework and explainability \\method with 60 perturbations.}
\label{tab:execution_times}
\renewcommand{\arraystretch}{1.3}
\scalebox{0.8}{
\begin{tabular}{ccccccc}
\hline \hline
\textbf{Framework} & \textbf{LIME} & \textbf{SMILE} & \textbf{Bay-LIME} \\ \hline
I-pix2pix & 1013.35 & \textbf{1208.08} & 1024.32\\ 
I2I-Turbo & 983.17 & \textbf{1222.09} & 980.12 \\
Diffuser-inpaint & 263.00 & \textbf{276.83} & 260.75 \\ 
\hline \hline
\end{tabular}
}
\end{table}

\subsubsection{Influence of Linguistic Elements on Explainability}
To understand the impact of different phrasing on the model's ability to interpret and execute image editing instructions, we designed an experiment using a structured set of prompts. Our aim is to evaluate how variations in wording affect the model’s performance and the clarity of the resulting edits. We selected ten diverse images and applied prompts such as ``Transform the image to depict a cloudy, dark evening,'' ``Transform the weather to snowy,'' and ``Turn the white-skinned person to a black-skinned person.'' For each control word in the prompts, we created variations using verbs, adjectives, nouns, ``ing'' forms, and rephrasing with ``make'' and ``must.'' Each image was processed using all prompt variations, and the outputs were analyzed using LIME to generate heatmaps showing the importance of each word.

\begin{figure}[H]
    \centering
    \includegraphics[width=0.8\linewidth]{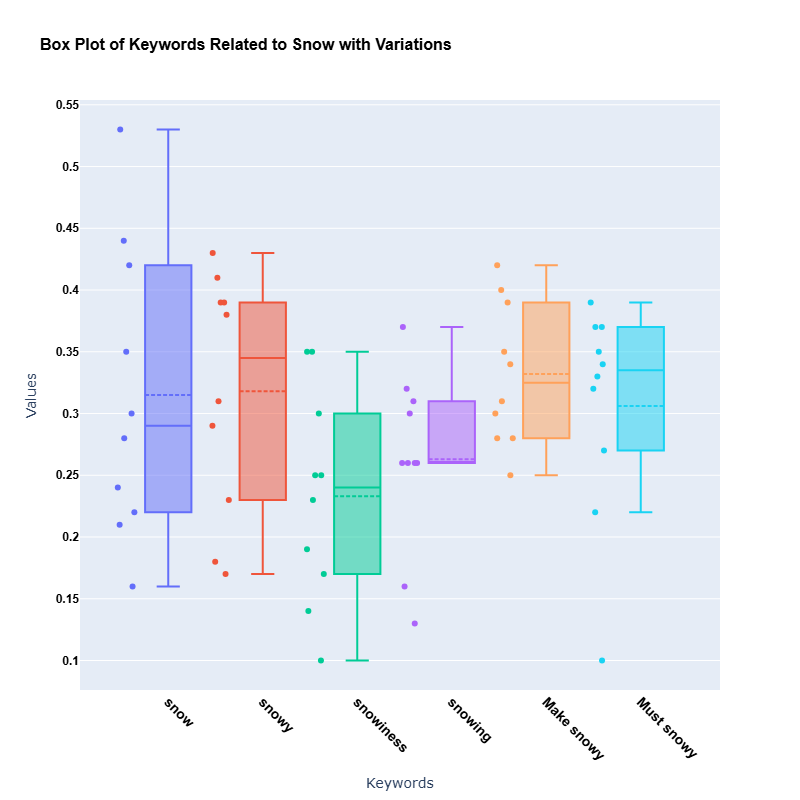}
    \caption{Effect of linguistic variations on responses to ``snow''\\ prompts. Different forms (e.g., noun, adjective, gerund) yield distinct output distributions.}
    \label{fig:fig_LE_S}
\end{figure}

Fig.~\ref{fig:fig_LE_S}, the snow-related prompts, reveals that noun and adjective forms lead to more consistent outputs, while gerund and directive phrases exhibit higher variability. This suggests that the model interprets straightforward descriptors more predictably, while transformation directives introduce flexibility and ambiguity in execution.

\begin{figure}[H]
    \centering
    \includegraphics[width=0.8\linewidth]{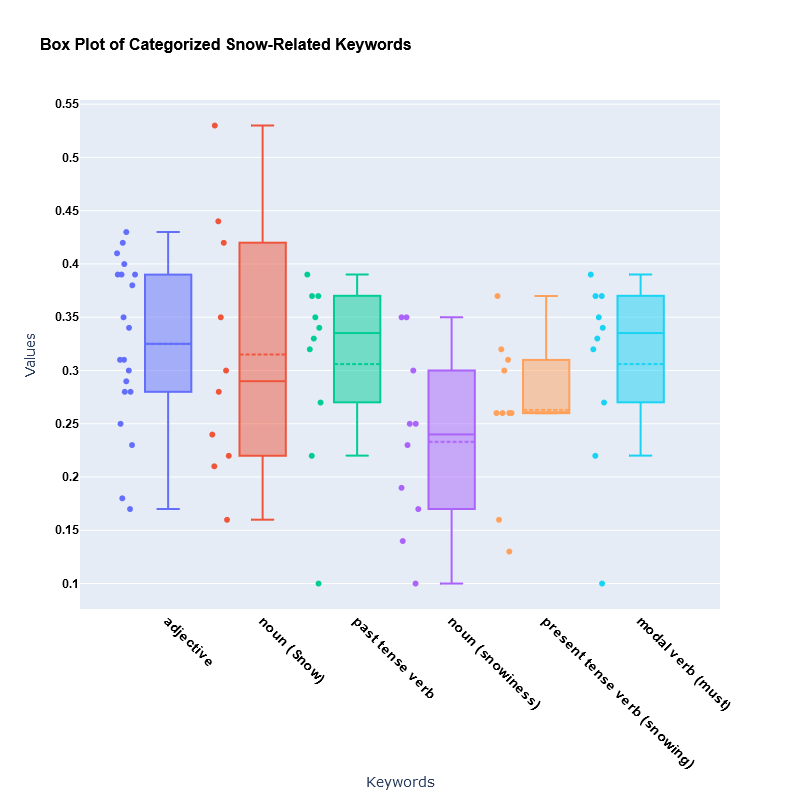}
    \caption{Response variations for a range of prompts across different words. This illustrates the model's general sensitivity to linguistic phrasing in instruction-based image editing.}
    \label{fig:fig_LE_G}
\end{figure}

Fig.~\ref{fig:fig_LE_G} shows the model's response to a range of prompts using various linguistic elements, providing a broader perspective on its sensitivity to phrasing. Testing directive phrases like ``make it'' reveals wider response ranges, suggesting that the model’s interpretation becomes more context-dependent with complex instructions.

\section{Conclusion}
Interpretability is a crucial aspect of machine learning and deep learning models. Human interaction with AI systems is essential for integrating these systems into the business, public, health, and education sectors. Since people are unlikely to adopt AI systems daily without understanding how decisions are made, developing interpretability solutions for these black box models is vital. Also, the extensive laws passed soon in AI will support the operation of a wide range of systems that could be more understandable to humans. Consequently, businesses that do not strive for interpretable models may face significant legal challenges in the future.
The text discusses the importance of enhancing transparency in diffusion-based image editing models, particularly concerning model interpretability. SMILE has been utilized to generate heatmaps that help illustrate how specific textual elements affect image edits, thereby clarifying the model’s decision-making process for users. This approach has been evaluated with various image editing models, including Instruct-Pix2Pix and Img2Img-Turbo, and has demonstrated reliable stability, accuracy, fidelity, and consistency. This framework has significant potential to build user trust and transparency, especially in critical fields such as healthcare and autonomous driving, where a clear understanding of model behavior is essential.

\section{Future Works}

\begin{enumerate}
\item Expanding to Other Generative Models: While our work has primarily focused on Instruct Image Editing models, future work could extend the principles and methodologies to other types of generative models, such as video generation. This could involve exploring both general-purpose video generation models, such as Sora~\cite{li2024sora}, and domain-specific models, like Gaia~\cite{hu2023gaia}.
\item Linking the prompt explainability with attention maps proposed by~\cite{tang2022daam}. By utilizing attention maps to visualize and interpret the model's response to specific prompt components, we could provide deeper insight into the relationship between prompt structure and generated output. This integration could help in building more transparent generative systems, as well as enhancing control over the generation process. Further exploration into how attention maps relate to semantic consistency across generated outputs could also improve model interpretability.
     
\end{enumerate}


\section*{CRediT authorship contribution statement}

\textbf{Zeinab Dehghani}: Conducted all aspects of the research, including writing the manuscript, performing tests, and overall project execution. \textbf{Koorosh Aslansefat}: Provided assistance with methodology, guidance and debugging through various stages of the research. \textbf{Adín Ramírez Rivera}: Offered valuable feedback and positive suggestions regarding the writing style. \textbf{Adil Khan, Franky George, Muhammad
Khalid} : Performed peer review.

\section*{Acknowledgements}
The author would like to express gratitude to \textbf{Ali Haghpanah Jahromi} from the University of Shiraz for his valuable insights and indirect contributions, which significantly influenced the quality of this work.

\section*{Data Availability}
The datasets used in this article are publicly available and codes and functions supporting this paper are published online at GitHub. For SMILE Explainability:
\\ \href{https://github.com/Dependable- Intelligent-Systems-Lab/xwhy}{https://github.com/Dependable-Intelligent-Systems-Lab/xwhy} \\ and for the proposed algorithm in this paper\\ \href{https://github.com/Sara068/Smile-to-Explain-Instruct Image-Editing}{https://github.com/Sara068/Smile-to-Explain-Instruct-Image-\\ Editing}.

\section*{Appendices}\label{sec14}

Table~\ref{tab:weather_transformations} provides a comprehensive table of transformation guidelines for visual and descriptive changes based on weather conditions. It serves as a vital resource for creators and designers who aim to produce authentic and compelling weather-themed visuals or narratives. This chart categorizes instructions by weather type (e.g., rainy, snowy, foggy) and offers specific prompt conditions, sentence structures, keywords, and example outputs for each. By following these structured guidelines, users can achieve precise and meaningful adaptations for various weather scenarios. The table provides a clear guide for modifying images to represent different weather scenarios, ensuring consistent and contextually accurate changes in visual outputs or descriptive narratives. 

\begin{table*}[h!]
    \centering
    \caption{Weather Condition Transformation and Corresponding Keywords}
    \label{tab:weather_transformations}
    \renewcommand{\arraystretch}{1.2}
    \begin{tabular}{>{\centering\arraybackslash}m{1.5cm}>{\centering\arraybackslash}m{3cm}>{\centering\arraybackslash}m{5cm}>{\centering\arraybackslash}m{3cm}>{\centering\arraybackslash}m{2.5cm}}
        \hline
        \hline
        \textbf{Condition} & \textbf{Specific Condition} & \textbf{Sentences} & \textbf{Keyword} & \textbf{Our Control} \\ 
        \cmidrule(lr){1-5}
        \multirow{15}{*}{Rainy} & Short Sentence & Transform the drizzle into a heavy downpour. & Transform, drizzle, heavy downpour & Heavy downpour \\
        \cmidrule(lr){2-5}
        & Short Sentence & Transform the image to a rainy scene with dark clouds, steady raindrops, and puddles forming on the ground. & Rainy scene, dark clouds, steady raindrops, puddles & Rainy scene \\
        \cmidrule(lr){2-5}
        & Short Sentence & Change the sunny weather to rainy, with dark clouds filling the sky and raindrops falling steadily. & Change, sunny weather, rainy, dark clouds, filling the sky, raindrops, falling steadily & Rainy, dark clouds, raindrops falling steadily \\
        \cmidrule(lr){2-5}
        & Length Sentence & Turn image to rainy and add a rainbow in the sky with a nice sense. & Turn, image, rainy, add, rainbow, sky, nice sense & Rainy, rainbow \\
        \cmidrule(lr){1-5}
        \multirow{8}{*}{Snowy} & Short Sentence & Transform the weather to snowy. & Transform, weather, snowy & Snowy \\
        \cmidrule(lr){2-5}
        & Short Sentence & Change the calm snow to heavy. & Change, calm snow, heavy & Heavy snow \\
        \cmidrule(lr){2-5}
        & Length Sentence & Change the bright day to a snowy scene, with snowflakes gently falling and the ground covered in a white blanket. & Change, bright day, snowy scene, snowflakes gently falling, ground covered, white blanket & Snowy scene, snowflakes, ground covered in a white blanket \\
        \cmidrule(lr){1-5}
        \multirow{10}{*}{Foggy} & Short Sentence & Change the foggy weather to a clear day. & Change, foggy weather, clear day & Clear day \\
        \cmidrule(lr){2-5}
        & Length Sentence & Transform the night image to depict a foggy evening where visibility of the road is poor. & Transform, night image, foggy evening, visibility of the road, poor & Foggy evening, poor visibility \\
        \cmidrule(lr){2-5}
        & Length Sentence & Switch the rain-drenched scene to one filled with fog, where the rain stops and a thick mist takes over, reducing visibility. & Switch, rain-drenched scene, filled with fog, rain stops, thick mist, reducing visibility & Filled with fog, thick mist, reducing visibility \\
        \cmidrule(lr){1-5}
        \multirow{10}{*}{Cloudy} & Short Sentence & Change the cloudy sky to a clear blue sky. & Change, cloudy sky, clear blue sky & Clear blue sky \\
        \cmidrule(lr){2-5}
        & Short Sentence & Transform the image to depict a cloudy, dark evening. & Transform, image, depict, cloudy, dark evening & Cloudy, dark evening \\
        \cmidrule(lr){2-5}
        & Length Sentence & Heavy clouds fill the sky with thunder rumbling and bright lightning bolts striking down, creating an intense and stormy atmosphere. & Heavy clouds, fill the sky, thunder rumbling, lightning bolts, striking down, intense stormy atmosphere & Intense, stormy atmosphere \\
        
        \hline
        \hline
    \end{tabular}
\end{table*}

Table~\ref{tab:table_of_co_2} presents a detailed table outlining transformation guidelines for visual and descriptive modifications based on time and human attributes, such as gender, age, and skin tone. This table is an essential resource for ensuring accurate and inclusive representations across diverse scenarios. This chart categorizes instructions by specific conditions and provides sentence structures, keywords, and example outputs for each attribute or time-related change. By offering clear examples and structured prompts, it helps users navigate the complexities of representing temporal and personal attributes effectively. It serves as a reference for adjusting images to reflect various temporal settings (e.g., day, night) and personal characteristics, ensuring consistent and accurate transformations that align with specified attributes or times of day. 

\begin{table*}
\centering
\caption{Table of Transformation Guidelines for Visual and Descriptive Changes Based on Time, and Human Attributes. This chart provides clear instructions for altering images or sentences to reflect different conditions, such as time of day, and personal characteristics like age, gender, and skin tone.}
\label{tab:table_of_co_2}
\setlength{\tabcolsep}{4pt}
\begin{tabular}{p{1.5cm}p{2.5cm}p{5cm}p{4cm}p{3cm}}
\hline
\hline
\textbf{Condition} & \textbf{Specific} & \textbf{Sentences} & \textbf{Keywords} & \textbf{Our Control} \\ \hline
\multirow{10}{*}{Person} 
& \multirow{4}{*}{Gender} 
    & Change the old woman to a young man in the image who is wearing sunglasses. 
    & Change, old woman, young man, wearing sunglasses 
    & Young man, wearing sunglasses \\ \cmidrule(lr){3-5}
& & Change the male in the image with a high-fashion woman. 
    & Change, male, high-fashion woman 
    & High-fashion woman \\ \cmidrule(lr){2-5}
& \multirow{4}{*}{Skin Color} 
    & Modify the person's complexion to a deep black hue. 
    & Modify, person's complexion, deep black hue 
    & Deep black hue \\ \cmidrule(lr){3-5}
& & Turn the white-skinned person to a black-skinned person. 
    & Turn, white-skinned person, black-skinned person 
    & Black-skinned person \\ \cmidrule(lr){2-5}
& \multirow{5}{*}{Age} 
    & Change the person's age to look considerably older. 
    & Change, person's age, considerably older 
    & Considerably older \\ \cmidrule(lr){3-5}
& & Change the elderly man into a young boy while retaining his distinctive facial characteristics. 
    & Change, elderly man, young boy, retaining, distinctive facial characteristics 
    & Young boy, distinctive facial characteristics \\ \cmidrule(lr){1-5}

\multirow{8}{*}{Tools} 
& \multirow{7.5}{*}{Replace} 
    & Remove the man’s hat in this foggy image. 
    & Remove, man’s hat, foggy image 
    & No hat \\ \cmidrule(lr){3-5}
& & Replace the traffic light with a simple light. 
    & Replace, traffic light, simple light 
    & Simple light \\ \cmidrule(lr){3-5}
& & Add a car crash to the street. 
    & Add, car crash, street 
    & Car crash \\ \cmidrule(lr){3-5}
& & Replace the black car with a white one. 
    & Replace, black car, white car 
    & White car \\ \cmidrule(lr){3-5}
& & Take off the man’s glasses and put them on the lady’s eyes. 
    & Take off, man’s glasses, put them on, lady’s eyes 
    & Glasses, the lady \\ \cmidrule(lr){2-5} 
& \multirow{1}{*}{change color} 
    & Convert the red bus to yellow one. 
    & Convert, red bus, yellow one 
    & Yellow  \\ \cmidrule(lr){3-5} \cmidrule(lr){1-5}

\multirow{10}{*}{Time} 

& \multirow{2}{*}{Day} 
    & Convert a nighttime scene into a bright, sunny daytime setting with clear blue skies. 
    & Convert, nighttime scene, bright, sunny daytime setting, clear blue skies 
    & Bright, sunny daytime, clear blue skies \\ \cmidrule(lr){3-5}
& & Alter the nighttime environment into a lively day, with the darkness lifting and the sunlight bringing the world to life. 
    & Alter, nighttime environment, lively day, darkness lifting, sunlight, bringing the world to life 
    & Lively day, sunlight \\ \cmidrule(lr){2-5}
& \multirow{3}{*}{Night} 
    & Convert it to night with some noise of lights. 
    & Nighttime, noise of lights 
    & Night, noise of lights \\ \cmidrule(lr){3-5}
& & Turn this beautiful image to night and turn on the lights. 
    & Nighttime, turn on lights 
    & Night, lights \\ \cmidrule(lr){3-5}
& & Turn it to night with snowy weather. 
    & Night, snowy weather 
    & Night, snow \\ 
\hline
\hline
\end{tabular}
\end{table*}

Table~\ref{fig:diff structure} provides a transformation matrix for sentence rewriting based on control words, illustrating methods to alter sentences by changing control words into various forms, such as verbs, adjectives, or nouns. This matrix serves as a practical guide for crafting dynamic and versatile sentence structures tailored to specific contexts or stylistic needs. By showcasing a variety of transformation examples, it helps users understand how subtle changes in control words can significantly impact sentence tone and meaning. This tool is especially valuable for content creators, linguists, and AI applications, offering a systematic approach to enhancing textual adaptability.

\begin{table*}[h!]
    \centering
    \caption{Transformation Matrix for Sentence Rewriting Based on Control Words: This table outlines various methods to modify sentences by transforming control words into different forms, such as verbs, adjectives, or nouns. It provides examples across diverse scenarios, including weather changes, visual alterations, and personal attributes, offering a structured approach to rephrasing for clarity and creativity.}
    \label{fig:diff structure}
    \renewcommand{\arraystretch}{1.2}
    \setlength{\tabcolsep}{3pt}
    \begin{tabular}{>{\centering\arraybackslash}m{2cm}>{\centering\arraybackslash}m{2cm}>{\centering\arraybackslash}m{2cm}>{\centering\arraybackslash}m{2cm}>{\centering\arraybackslash}m{2.5cm}>{\centering\arraybackslash}m{2.5cm}>{\centering\arraybackslash}m{2.5cm}}
        \hline
        \hline
        \textbf{Control Words} & Cloudy & Dark & Snowy & Rain, Rainbow & Heavy Downpour & Black-skinned person \\ 
        \hline
        \textbf{Control Words as Verb} & Transform the image to be clouded and darkened. & Transform the image to be clouded and darkened. & Transform the weather to snow. & Turn the image to raining and add a rainbow in the sky with a nice sense. & Transform the drizzle to downpour heavily. & Turn the white-skinned person to blacken. \\ 
        \hline
        \textbf{Control Words as Adjective} & Transform the image to depict a cloudy, dark evening. & Transform the image to depict a darkening, overcast evening. & Transform the weather to snowy. & Turn the image to rainy and add a rainbow-colored sky with a nice sense. & Transform the drizzle to make it heavily downpouring. & Turn the person to make them black-skinned. \\ 
        \hline
        \textbf{Control Words as Noun} & Transform the image to depict the dark of a cloudy evening. & Transform the image to depict the dark of a cloudy evening. & Transform the weather to snowiness. & Turn the image to rain and add a rainbow in the sky with a nice sense. & Transform the drizzle into a downpouring rain. & Turn the white-skinned person into a person with black skin. \\ 
        \hline
        \textbf{Use Word with 'ing'} & The image is depicting a cloudy, dark evening. & The image is depicting a cloudy, dark evening. & Transform the weather to snowing. & Turn the image to rainy and add a rainbowing sky with a nice sense. & Transform the drizzle into a heavy downpouring. & Turn the white-skinned person into a person with black-skinning. \\ 
        \hline
        \textbf{Sentences with Make} & Make the image depict a cloudy, dark evening. & Make the image depict a cloudy, dark evening. & Make the weather to snowy. & Make the image to rainy and add a rainbow in the sky with a nice sense. & Make the drizzle into a heavy downpour. & Make the white-skinned person into a person with black skinning. \\ 
        \hline
        \textbf{Sentences with Must} & The image must depict a cloudy, dark evening. & The image must depict a cloudy, dark evening. & The weather must be snowy. & The image must be turned to rainy, and a rainbow must be added in the sky with a nice sense. & The drizzle must be transformed into a heavy downpour. & The white-skinned person must be turned into a black-skinned person. \\ 
        \hline
        \hline
    \end{tabular}
\end{table*}

\begin{figure*}
    \centering
    \includegraphics[width=1\linewidth]{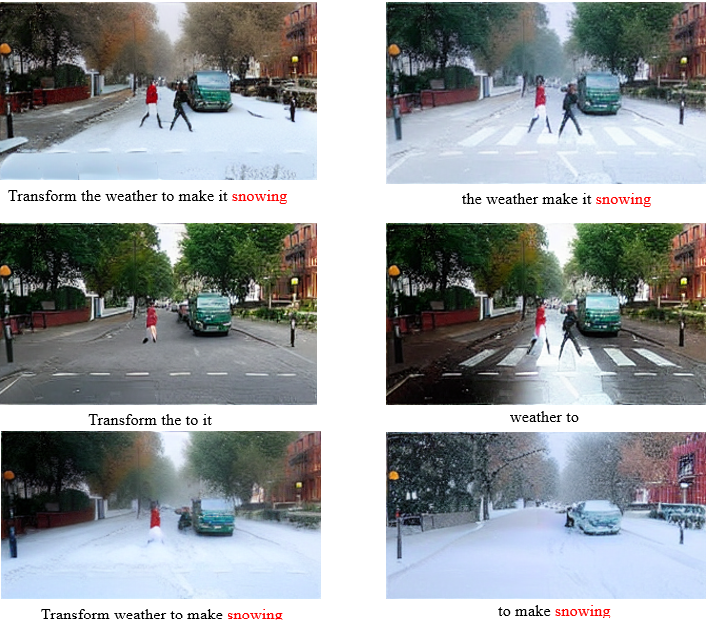}
    \caption{Results generated by Pix2Pix showing how the presence of the keyword ``snowing'' in the perturbation prompts leads to snowy scenes, while its absence results in non-snowy images, illustrating the model's responsiveness to specific keywords.}
    \label{fig:Snowing_per}
\end{figure*}

In our model, we refined input processing by splitting the text into individual words and creating targeted perturbations based on these words. This approach enhances granularity in understanding how each keyword influences the output. Fig.~\ref{fig:Snowing_per} illustrates how different models respond to prompts containing specific keywords. For instance, with the keyword “snowing” included, the generated image accurately depicts a snowy scene. However, when the keyword is omitted, the image reflects alternative visual outcomes. This highlights the model’s sensitivity to precise textual cues and the importance of keyword selection in achieving desired results.

This method enables us to assess the direct impact of individual keywords on image generation, revealing how each word shapes the output. By identifying and quantifying these effects, we can refine the model to enhance consistency and accuracy in generating context-specific visuals. By examining these effects, we can further optimize the model to ensure that subtle changes in textual input lead to precise and predictable alterations in the generated images. This methodology is a key step toward achieving robust and reliable AI-generated content.

\begin{figure*}[tb]
    \centering
    \includegraphics[width=1\linewidth]{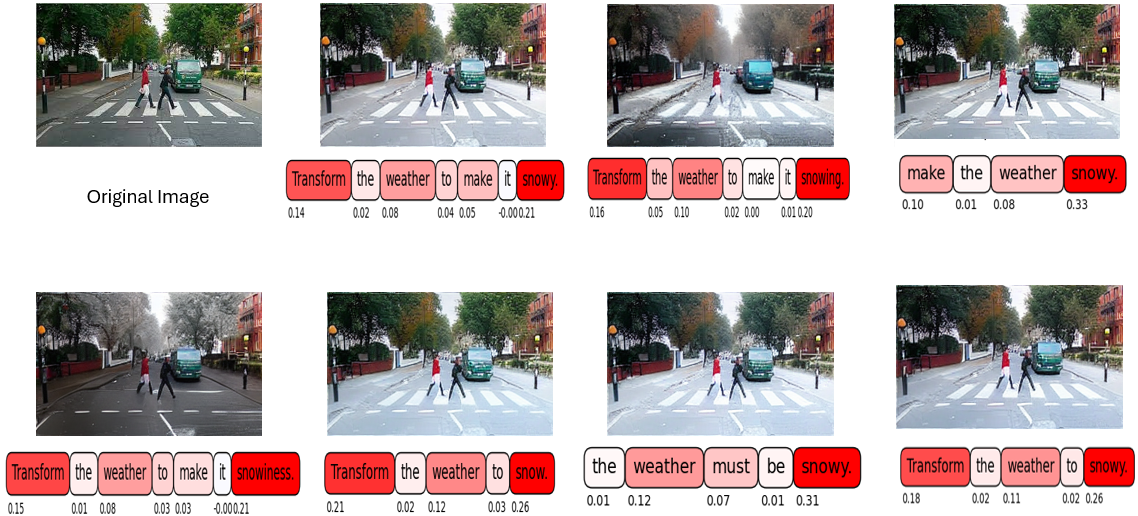}
    \caption{Comparison of different structures of prompts used to generate the same image using Pix2Pix, showcasing how variations in the prompt's structure lead to distinct transformations of the same base scenario.}
    \label{fig:Snow_str}
\end{figure*}

Fig.~\ref{fig:Snow_str} shows how an image changes when different prompt structures are used. This visualization underscores the pivotal role of prompt design in guiding AI-generated outputs. By tweaking keywords and phrases like “make it snowy” or “the weather must be snowy,” we see noticeable differences in how the image looks. Such variations illustrate the importance of nuanced prompt crafting to achieve specific visual goals. The importance of each word in the prompts is reflected by its weight, with darker colors representing words that have a more substantial influence. This weighted representation provides valuable insights into the linguistic components that drive visual transformations.

This analysis demonstrates the model's sensitivity to specific prompt elements and highlights how prompt phrasing directly impacts the generated image's characteristics. By leveraging this understanding, users can refine their prompts to produce more accurate and contextually relevant outputs, making this approach a powerful tool for optimizing AI-driven image generation.

\begin{figure*}
    \centering
    \includegraphics[width=1\linewidth]{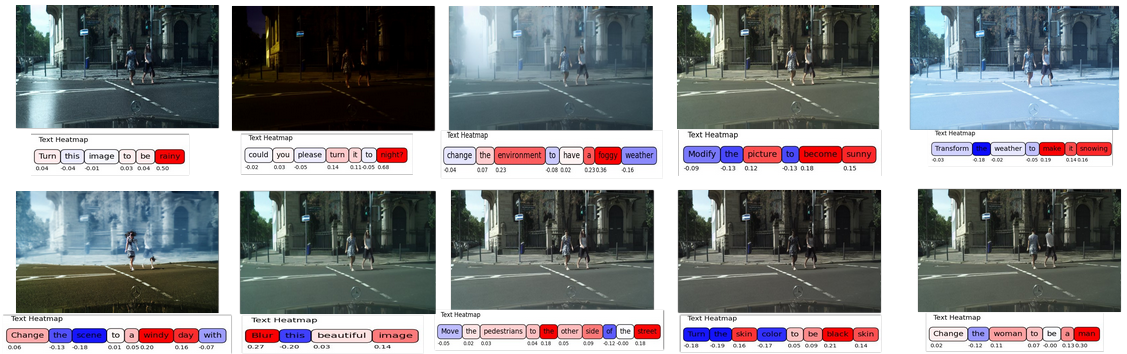}
    \caption{Heatmap analysis of prompts displaying the weights of individual words in guiding the Pix2Pix model’s visual output. Each row shows transformations generated from different prompt structures, with heatmaps illustrating how specific words affect the model’s interpretation and the resulting image.}
    \label{fig:diff_prompt}
\end{figure*}

Fig.~\ref{fig:diff_prompt} presents the heatmap analysis of prompt words to illustrate the influence of each word on the generated images using the Pix2Pix model. This analysis provides a visual breakdown of how the model decodes and assigns importance to textual inputs. Each image corresponds to a specific prompt, with heatmaps showing the weight assigned to each word. Darker colors represent words with higher influence, indicating their more substantial impact on the resulting image transformation. This allows users to pinpoint which words are most critical for achieving desired visual outcomes.

This analysis highlights how the model interprets and prioritizes individual words in the prompt, allowing us to observe the relationship between textual emphasis and visual modifications. Such insights are invaluable for refining prompt engineering strategies, ensuring more precise and predictable image transformations. By understanding these dynamics, we can further optimize the Pix2Pix model for enhanced sensitivity to nuanced textual cues, paving the way for improved AI-generated content.

\begin{figure*}
    \centering
    \includegraphics[width=1\linewidth]{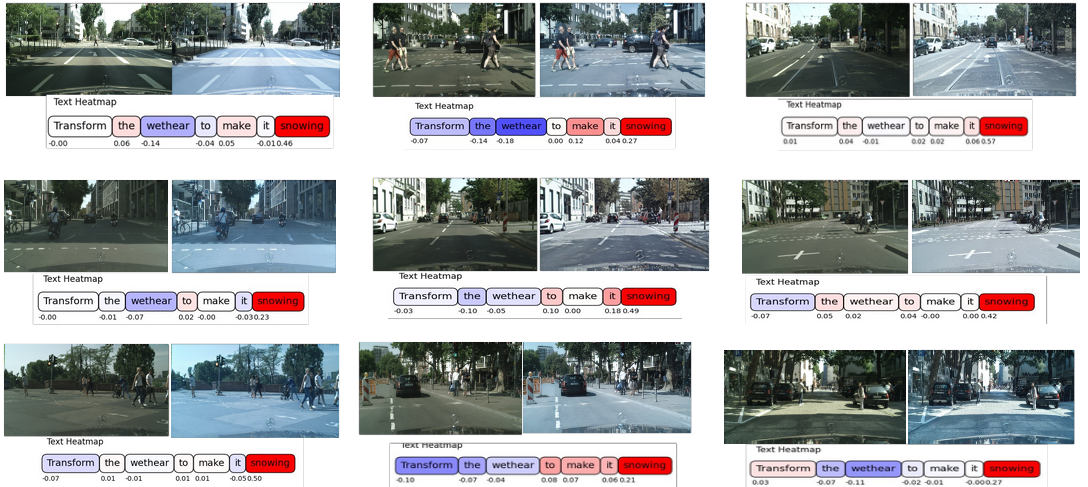}
    \caption{ Heatmap analysis of a single prompt applied across different images. Each heatmap shows the weight of each word in the prompt, indicating how the Pix2Pix model maintains consistent word influence across varied scenes to achieve the desired transformations.}
    \label{fig:diff_images}
\end{figure*}

\begin{figure*}
    \centering
    \includegraphics[width=1\linewidth]{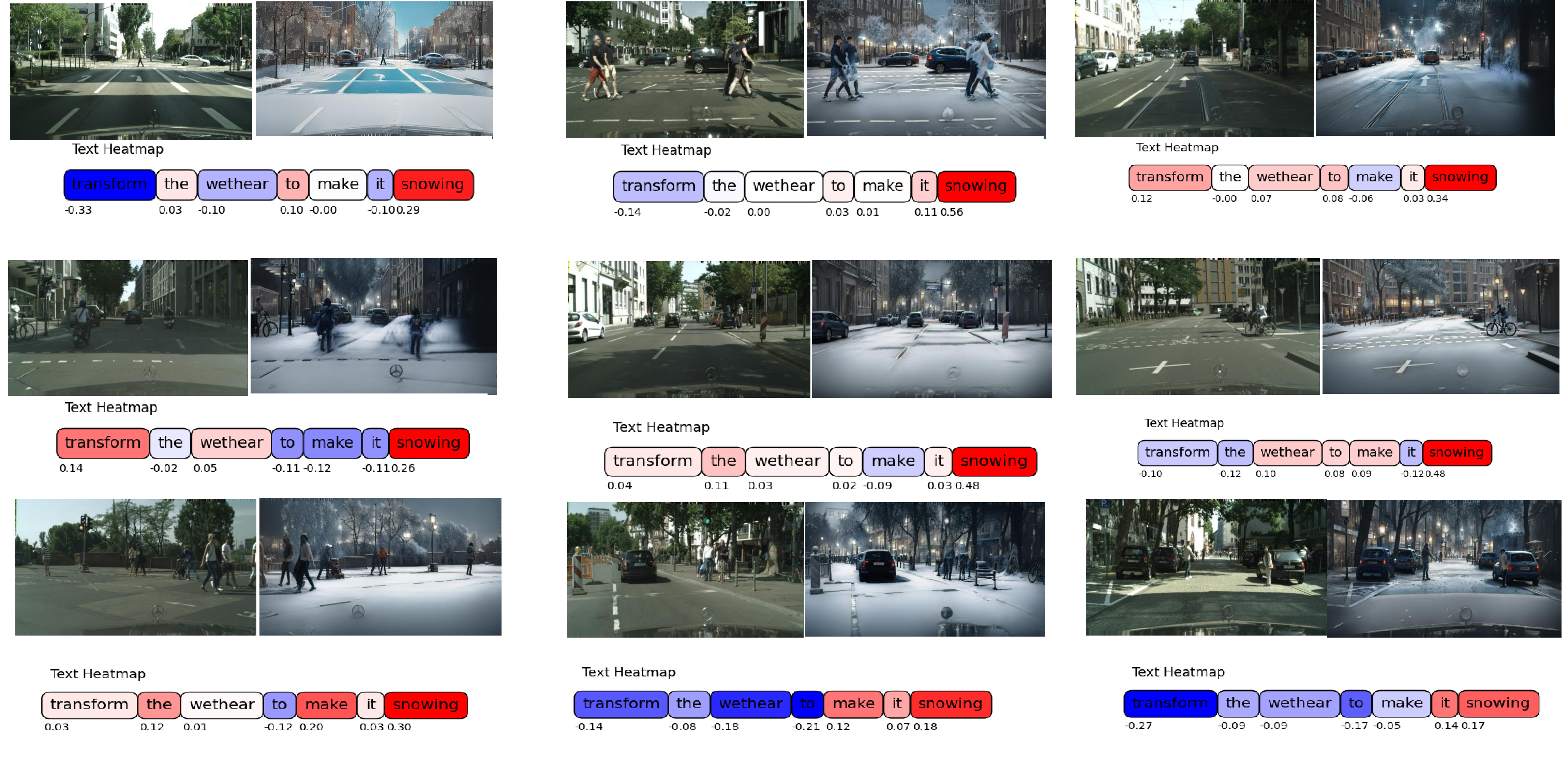}
    \caption{ Heatmap analysis of a single prompt applied across different images. Each heatmap shows the weight of each word in the prompt, indicating how the I2I-turbo model maintains consistent word influence across varied scenes to achieve the desired transformations.}
    \label{fig:diff_images_turbo}
\end{figure*}

Fig.~\ref{fig:diff_images},~\ref{fig:diff_images_turbo}illustrates the consistency of word influence in a single prompt applied across different images. This demonstrates the robustness of the models in maintaining predictable word-driven transformations across varying scenarios. The heatmaps display the weights assigned to each word in the prompt, with darker colors signifying higher influence. Such visual representation helps in understanding how individual words dominate the transformation process. The keyword “snowing” has a strong impact, resulting in reliable visual changes that align with the modification. This highlights the effectiveness of the keyword in directing the model’s focus toward the desired weather condition.

This analysis shows how the Pix2Pix and I2I-turbo models interpret the same prompt across various contexts.


\end{document}